\documentclass[lettersize,journal]{IEEEtran}
\usepackage{amsmath,amsfonts}
\usepackage{algorithmic}
\usepackage{algorithm}
\usepackage{array}
\usepackage[caption=false,font=normalsize,labelfont=sf,textfont=sf]{subfig}
\usepackage{textcomp}
\usepackage{stfloats}
\usepackage{url}
\usepackage{verbatim}
\usepackage{graphicx}
\usepackage{cite}
\usepackage{color}
\usepackage{booktabs}
\usepackage{bm}
\usepackage{amssymb}
\usepackage{multirow} 
\usepackage{makecell}
\usepackage{footnote}
\usepackage{xcolor}
\usepackage{arydshln}
\usepackage{times}  
\usepackage{helvet}  
\usepackage{courier}  
\begin{document}

\title{Stable-Hair v2: Real-World Hair Transfer via Multiple-View Diffusion Model}

\author{Kuiyuan~Sun*,
        Yuxuan~Zhang*,
        Jichao~Zhang*,
        Jiaming~Liu,
        Wei~Wang,~\IEEEmembership{Member,~IEEE},
        Nicu~Sebe,~\IEEEmembership{Senior~Member,~IEEE},
        and Yao Zhao,~\IEEEmembership{Fellow,~IEEE},
\IEEEcompsocitemizethanks{\IEEEcompsocthanksitem Kuiyuan Sun, Wei Wang, Yao Zhao are with the Institute of Information Science, Beijing Jiaotong University, Beijing, China. E-mail: 20112001@bjtu.edu.cn, wei.wang@bjtu.edu.cn, yzhao@bjtu.edu.cn.
\IEEEcompsocthanksitem Jichao Zhang is with the School of Computer Science, Ocean University of China, Qingdao, China. E-mail: zhang163220@gmail.com.
\IEEEcompsocthanksitem Yuxuan Zhang is with the Department of Computer Science and Engineering, The Chinese University of Hong Kong, Hong Kong, China. E-mail: yxzhang@cse.cuhk.edu.hk.
\IEEEcompsocthanksitem Jiaming Liu is with the Artificial Intelligence department, Tiamat AI, Shanghai, China. E-mail: jmliu1217@gmail.com.
\IEEEcompsocthanksitem Nicu Sebe is with the Department of Information Engineering and Computer Science (DISI), University of Trento, Italy. E-mail: sebe@disi.unitn.it.}
\thanks{The first three authors contribute equally; Wei Wang and Yao Zhao are the corresponding authors.}}



\markboth{}%
{Shell \MakeLowercase{\textit{et al.}}: A Sample Article Using IEEEtran.cls for IEEE Journals}

\IEEEpubid{}

\maketitle

\begin{figure*}[!t]
    \centering
    \includegraphics[width=1.00\linewidth]{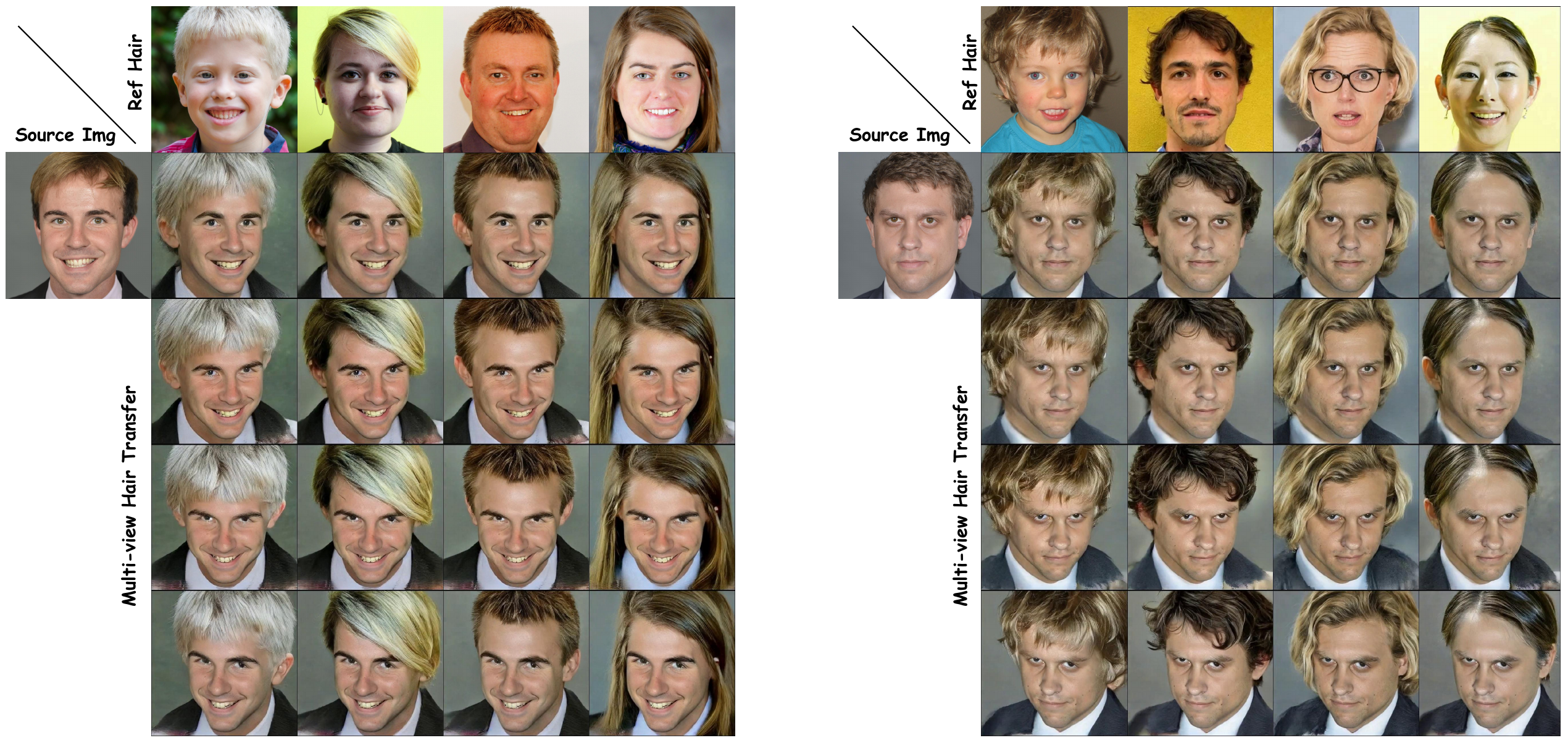}
    \vspace{-0.25cm}
    \caption{Given a source and a reference image, Stable-Hair v2 precisely transfers key hair attributes—color, length, curvature, and overall structure—onto the source while faithfully preserving the source’s identity. Built on a multi-view diffusion framework, our method generates photorealistic results that remain seamless and temporally consistent across arbitrary viewpoints.}
    \vspace{-0.3cm}
    \label{fig:teaser}
\end{figure*}

\begin{abstract}

While diffusion-based methods have shown impressive capabilities in capturing diverse and complex hairstyles, their ability to generate consistent and high-quality multi-view outputs—crucial for real-world applications such as digital humans and virtual avatars—remains underexplored. In this paper, we propose Stable-Hair v2, a novel diffusion-based multi-view hair transfer framework. To the best of our knowledge, this is the first work to leverage multiple-view diffusion models for robust, high-fidelity, and view-consistent hair transfer across multiple perspectives. We introduce a comprehensive multi-view training data generation pipeline comprising a Diffusion-based Bald Converter, a Data-Augment Inpainting Model, and a Face-Finetuned Multi-View Diffusion Model to generate high-quality triplet data, including bald images, reference hairstyles, and view-aligned source-bald pairs. Our multi-view hair transfer model integrates polar-azimuth embeddings for pose conditioning and temporal attention layers to ensure smooth transitions between views. To optimize this model, we design a novel multi-stage training strategy consisting of Pose-Controllable Latent IdentityNet training, Hair Extractor training, and Temporal Attention training. Extensive experiments demonstrate that our method accurately transfers detailed and realistic hairstyles to source subjects while achieving seamless and consistent results across views, significantly outperforming existing methods and establishing a new benchmark in multi-view hair transfer. Code is publicly available at https://github.com/sunkymepro/StableHairV2.

\end{abstract}

\begin{IEEEkeywords}
Multiple-View Diffusion Models, Avatar Generation, Hair Transfer.
\end{IEEEkeywords}

\section{Introduction}

Hair transfer is one of the most challenging tasks in the virtual try-on domain. The objective of this task is to transfer hair color, shape, and structure attributes from a reference image to a user-provided source image while preserving the identity of the source image. While GAN-based approaches~\cite{tan2020michigan,2018arXivhair-Gans,2022CtrlHair,2022HairNet,2023HairNeRF,chung2022hairfit,zhu2021barbershop,saha2021LOHO,wei2022hairclip,wei2023hairclipv2,nikolaev2024hairfastgan,Khwanmuang2023StyleGANSalon,kim2022styleSYH} have advanced this field, they often struggle with the complexity and diversity of real-world hairstyles. The emergence of diffusion models has revolutionized image generation, offering superior training stability and output quality. Recent diffusion-based attempts~\cite{zhang2025stable,chung2025hairfusion} have demonstrated remarkable accuracy in hair transfer and identity preservation. Beyond these achievements, diffusion models have shown exceptional capabilities in multi-view 3D object generation. This inspires us to explore a novel direction: \textit{Can we leverage diffusion models to generate multi-view consistent hair transfer results?}

To achieve multi-view hair transfer, two key challenges must be addressed: (1) acquiring a multi-view hair dataset, and (2) enabling the hair transfer model to generate consistent results across different viewpoints.

To tackle the first challenge, we propose a \textbf{multi-view training data generation pipeline} comprising three key steps:

\begin{enumerate}
    \item \textbf{Diffusion-based Bald Converter}: Following StableHair~\cite{zhang2025stable}, we adopt a diffusion-based converter to generate bald proxy images from user-provided source images, achieving superior identity and background preservation compared to prior GAN-based approaches~\cite{Wu_2022_CVPR}.
    
    \item \textbf{Data-Augmented Inpainting Model}: We employ an inpainting model to generate diverse reference images that share the same hairstyle but vary in identity and background. We observe that the model often retains irrelevant information (e.g., background color) from the source image, potentially misleading the hair extractor to learn non-hair-related features. To mitigate this, we propose a data augmentation strategy that uses a face-and-hair mask to isolate key regions and combines them with diverse backgrounds generated by Stable Diffusion. This encourages the model to focus more accurately on hairstyle features.
    
    \item \textbf{Face-Finetuned Multi-View Diffusion Model}: Due to the lack of open-source multi-view hair datasets, we leverage a pretrained large-scale multi-view diffusion model to synthesize view-consistent image pairs for both source and bald inputs. While this model serves as a general-purpose solution for multi-view generation, it exhibits limitations in view reconstruction accuracy and controllability when applied to human faces. To address this, we introduce the \textbf{Face-Finetuned Multi-View Diffusion Model}, which fine-tunes the base model on facial data to enhance its capability in generating coherent and identity-preserving multi-view facial images.
\end{enumerate}

Addressing the second challenge requires the hair transfer model to synthesize images from specified viewpoints while ensuring smooth transitions across consecutive views. Inspired by recent advances in multi-view diffusion models~\cite{voleti2024sv3d}, we propose the Pose-Controllable Latent IdentityNet. The term Latent highlights the model’s ability to preserve the identity and background of the source image throughout the transfer process, while Pose-Controllable refers to the incorporation of polar and azimuth angles into an augmented embedding. This pose embedding is fused with the diffusion model's time embedding, enabling control over the viewing direction of the generated images.

However, we observe that jointly training the pose control and hair transfer tasks leads to insufficient learning of pose control, resulting in inconsistent head shapes across different views. To address this issue, we adopt a stage-wise training strategy: we first train the model to learn pose alignment using bald multi-view images, which avoids interference from hair details and promotes better shape consistency. After pose control is established, we train the hairstyle transfer module to synthesize hair conditioned on the aligned head shapes. To further improve temporal coherence in multi-view hair transfer, we introduce a temporal attention layer into the U-Net blocks of the diffusion model. This layer performs self-attention across consecutive frames, allowing the model to capture temporal dependencies and generate consistent and natural motion of both head and hair across viewpoints.

In summary, our method leverages a multi-stage training pipeline consisting of three core components: (1) the Pose-Controllable Latent IdentityNet for viewpoint-consistent identity preservation, (2) a Hair Extractor module for accurate hairstyle synthesis, and (3) a Temporal Attention mechanism for inter-frame consistency. Together, these components significantly improve the quality and coherence of multi-view hair transfer results. Through extensive experiments, Stable-Hair v2 has demonstrated its superior performance, significantly surpassing existing state-of-the-art hair transfer methods in terms of fidelity, fine-grained detail, and multi-view consistency. 



In summary, our contributions are: 


\begin{itemize}
    \item In this paper, we introduce \textit{Stable-Hair v2}, the first comprehensive framework for multi-view hairstyle transfer. Extensive experimental evaluations demonstrate that our approach not only enables efficient one-step multi-view hairstyle synthesis, but also establishes new state-of-the-art performance in both photorealistic fidelity and fine-grained detail preservation, outperforming existing methods by a significant margin. 
    \item We design a novel training data generation pipeline for constructing multi-view triplets (\textit{Source}, \textit{Reference}, \textit{Bald}) to support the training of multi-view hairstyle transfer models. This pipeline integrates three key components: a diffusion-based bald converter, a data-augmented inpainting model, and a face-finetuned multi-view diffusion model.
    
    \item We develop a diffusion-based multi-view hair transfer model that extends traditional single-view generation to consistent and controllable multi-view synthesis. Additionally, we introduce a multi-stage training strategy to further enhance hair transfer quality and improve view consistency across generated results.
\end{itemize}


\section{Related Works}

\subsection{Hair Style Transfer}
The rapid development of GAN-based methods~\cite{tan2020michigan,2018arXivhair-Gans,2022CtrlHair,2022HairNet,2023HairNeRF,chung2022hairfit,zhu2021barbershop,saha2021LOHO,wei2022hairclip,wei2023hairclipv2,nikolaev2024hairfastgan,Khwanmuang2023StyleGANSalon,kim2022styleSYH,shu2022few} has significantly advanced the field of hairstyle transfer. Most existing approaches rely on generative adversarial networks (GANs), with various strategies developed to enhance control, fidelity, and realism.

\begin{figure*}[h]
    \centering
 \includegraphics[width=1.0\linewidth]{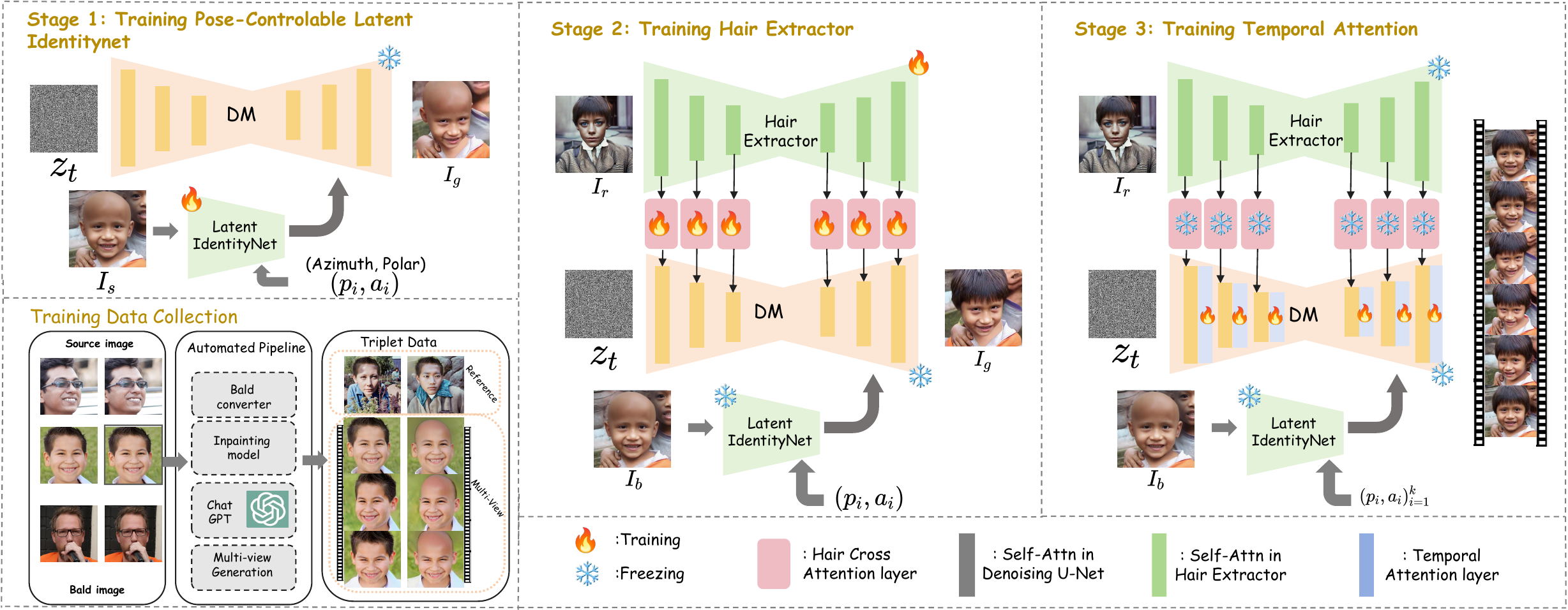}
    \caption{The proposed multi-view hair transfer model employs a multiple stage training strategy: (1) training a Pose-Controllable Latent IdentityNet with multi-view data for view control; (2) training a Hair Extractor Module using a triplet dataset to transfer detailed reference hair onto a bald proxy image; and (3) incorporating and training a Temporal Attention Module within the UNet blocks to ensure view-consistent outputs. And our triplet dataset is collected by the proposed pipeline (Bottom-Left).}
    \vspace{-0.3cm}
    \label{fig:method}
\end{figure*}

MichiGAN~\cite{tan2020michigan} decomposes hairstyles into four orthogonal attributes and designs dedicated modules for representation, manipulation, and recombination, forming an end-to-end architecture. Barbershop~\cite{zhu2021barbershop} introduces a novel latent space blending technique that excels in preserving details and spatial cues by extracting hairstyle features from multiple reference images. LOHO~\cite{saha2021LOHO} proposes an optimization-based method using GAN inversion to recover detailed hair structures in the latent space, leveraging a two-stage optimization scheme with gradient orthogonalization for disentangled attribute editing.

Hairmapper~\cite{Wu_2022_CVPR} trains a dedicated network to remove hair within StyleGAN’s latent space, serving as a pre-processing step for hair transfer. While these methods are effective for basic hairstyle transfers, they generally fail to handle large pose variations. To address this, SYH~\cite{kim2022styleSYH} introduces a pose-invariant model capable of maintaining local texture under significant viewpoint changes. HairCLIP~\cite{wei2022hairclip} utilizes CLIP-based guidance for unified hair editing, and HairCLIPV2~\cite{wei2023hairclipv2} reformulates hair editing as transfer tasks involving diverse proxies. StyleGAN-Salon~\cite{Khwanmuang2023StyleGANSalon} improves transfer accuracy through multi-view optimization and guided supervision. HairNeRF~\cite{2023HairNeRF} integrates 3D-aware GANs with NeRF-like components to enable multi-view hair synthesis, combining modules for inversion, alignment, recombination, and blending. HairFastGAN~\cite{nikolaev2024hairfastgan} proposes an efficient encoder to support fast hairstyle transfer even under varying poses.

Despite these advances, GAN-based methods still struggle with complex hairstyle variations in real-world scenarios, especially under unconstrained conditions. To address these limitations, our earlier work~\cite{zhang2025stable} was the first to introduce a diffusion-based framework for high-quality and robust hairstyle transfer. Concurrent approaches~\cite{chung2025preserve} also explore diffusion models for in-the-wild hair editing. However, most prior works remain constrained to single-view settings, limiting their applicability.

In this work, we extend the diffusion-based approach to a multi-view framework that achieves high-quality, robust, and view-consistent hairstyle transfer, significantly advancing the practicality and controllability of real-world applications.

\subsection{Diffusion Models}
Diffusion models, especially Denoising Diffusion Probabilistic Models (DDPM)~\cite{ho2020denoising}, are a class of generative models that learn data distributions through a two-stage process: a forward process that gradually adds noise to the data until it becomes a pure Gaussian distribution, and a reverse process that iteratively denoises samples from this distribution to reconstruct data consistent with the original training distribution. Currently, diffusion models have attracted a great deal of attention and have seen significant advancements. As the most prominent generative models today, diffusion models have achieved state-of-the-art results across various image generation tasks, including text-to-image generation~\cite{dalle2, sdxl, Imagen, rombach2022high, IF}, image editing~\cite{attendandexcite, layerdiffusion, selfguidance, masactrl,dragondiffusion,iedit,sine,tsaban2023ledits,li2024zone}, controllable generation~\cite{controlnet,t2i,unicontrol,directed}, personalized image generation~\cite{zhang2024ssr, lora, domainagnostic,taming,TI, hyperdreambooth, DB,zhang2024fast} and so on. In particular, the recent emergence of diffusion-based virtual try-on applications further demonstrates the formidable generative capabilities of diffusion models, enabling commercial grade virtual try-ons~\cite{xu2024ootdiffusion,stablegarment,kim2023stableviton,zeng2023cat} and virtual makeovers~\cite{zhang2024stable}, which were previously unattainable with traditional GAN methods. Diffusion models excel in generating high-quality images, especially when trained on 2D datasets.  In this paper, We use diffusion models to achieve high-fidelity and robust hairstyle transfers, addressing the limitations of GAN-based methods and significantly enhancing the quality and realism of complex hairstyles.

Diffusion models have also been widely adopted for 3D object generation. Compared to 3D reconstruction methods~\cite{peng2023implicit,peng2021neural,qian2024gaussianavatars} based on NeRF~\cite{mildenhall2020nerf} or 3D Gaussian representations~\cite{kerbl3Dgaussians}, generation-based techniques can significantly accelerate 3D asset creation and eliminate the need for complex reconstruction pipelines. Given a single image of an object, these methods~\cite{liu2023zero,shi2023mvdream,shi2023zero123++,liu2023syncdreamer} use a diffusion model to synthesize novel views conditioned on the input image and target camera poses. Specifically, ZERO-123 introduce a view-conditioned latent diffusion model with encoding the relative camera extrinsics into the conditional embedding. Additionally, Zero-123~\cite{liu2023zero} pre-trains the model using synthetic 3D datasets to learn rich 3D priors, i.e., sampling from Objaverse~\cite{deitke2023objaverse}. However, a major challenge for diffusion-based 3D generation is maintaining view consistency across different viewpoints. To address this, SV3D~\cite{voleti2024sv3d} incorporates temporal attention mechanisms into the U-Net architecture, inspiring by video generation models~\cite{blattmann2023stable}. In this paper, we are the first to apply a multi-view diffusion model to the hair transfer task, and we introduce a multi-stage training strategy to achieve high-quality, view-consistent hair transfer results.

\section{Methodology}

We present Stable-Hair v2, a diffusion-based framework designed for multi-view hairstyle transfer. As illustrated in Fig.\ref{fig:method} and Fig.\ref{fig:data}, Stable-Hair v2 consists of a Multi-View Training Data Generation Pipeline and a Multi-View Hair Transfer Model, enabling the transfer of diverse hairstyles from reference images to a source image while generating consistent multi-view results. Our approach leverages the latent diffusion model Stable Diffusion~\cite{rombach2022high} as the foundational backbone.

\subsection{Latent Diffusion Model}

Unlike the original diffusion model, the latent diffusion model (LDM) incorporates an autoencoder consisting of an encoder–decoder pair. The encoder compresses an input image into a compact, high-level latent representation, which the decoder subsequently reconstructs into the full-resolution image space. Consequently, the UNet that governs the diffusion process operates in latent space rather than directly on pixel space. Under this formulation, the training objective of LDM can be reformulated as:

\begin{equation} 
\mathcal{L}_{mse} = \mathbb{E}_{z_{t}, t, \epsilon, \mathbf{c}}(|| \epsilon - \epsilon_{\theta}(z_{t}, t, \mathbf{c})||),
\end{equation}
where $\epsilon_{\theta}$ represents the forward process of UNet in LDM, $z_t$ is the noisy latent $z$ under timestep $t$, and $\mathbf{c}$ is the conditional information in Stable Diffusion. 

\subsection{Multi-View Training Data Generation Pipeline} 

We propose a novel pipeline for constructing training triplets consisting of a source image $I_s$, a reference image $I_r$, and a corresponding bald image $I_b$. Simultaneously, the pipeline generates multi-view image pairs $\{(I_s^i, I_b^i)\}_{i=1}^{K}$, where $K$ denotes the number of views, ensuring consistency across both original and bald representations.

\subsubsection{Diffusion-Based Bald Converter} Instead of relying on GAN-based methods~\cite{Wu_2022_CVPR}, which often suffer from low-quality hair removal and identity distortion, we follow StableHair~\cite{zhang2025stable} to utilize a diffusion-based approach to convert source images $I_{s}$ into their bald counterparts $I_{b}$, as shown in Fig.~\ref{fig:data} (Step 1). Specifically, we leverage a vanilla diffusion model combined with our custom-designed Latent ControlNet architecture (detailed in the following section) to perform this transformation.

Once trained, our Bald Converter can effectively remove hair from $I_{s}$ without the need for image alignment or cropping. This results in cleaner, more consistent bald images, providing a robust and uniform baseline for subsequent hairstyle transfer. By standardizing the input to a bald state, this method significantly enhances visual fidelity and realism in the generated outputs. Importantly, this standardization step plays a critical role in improving hairstyle transfer performance. By removing hair-related variability, the model can focus more precisely on facial features and identity-preserving details. This not only increases training stability but also leads to higher accuracy and consistency in hairstyle generation.

\subsubsection{Data-Augmented Inpainting Model} 
\begin{figure}[!t]
    \centering
    \includegraphics[width=1\linewidth]{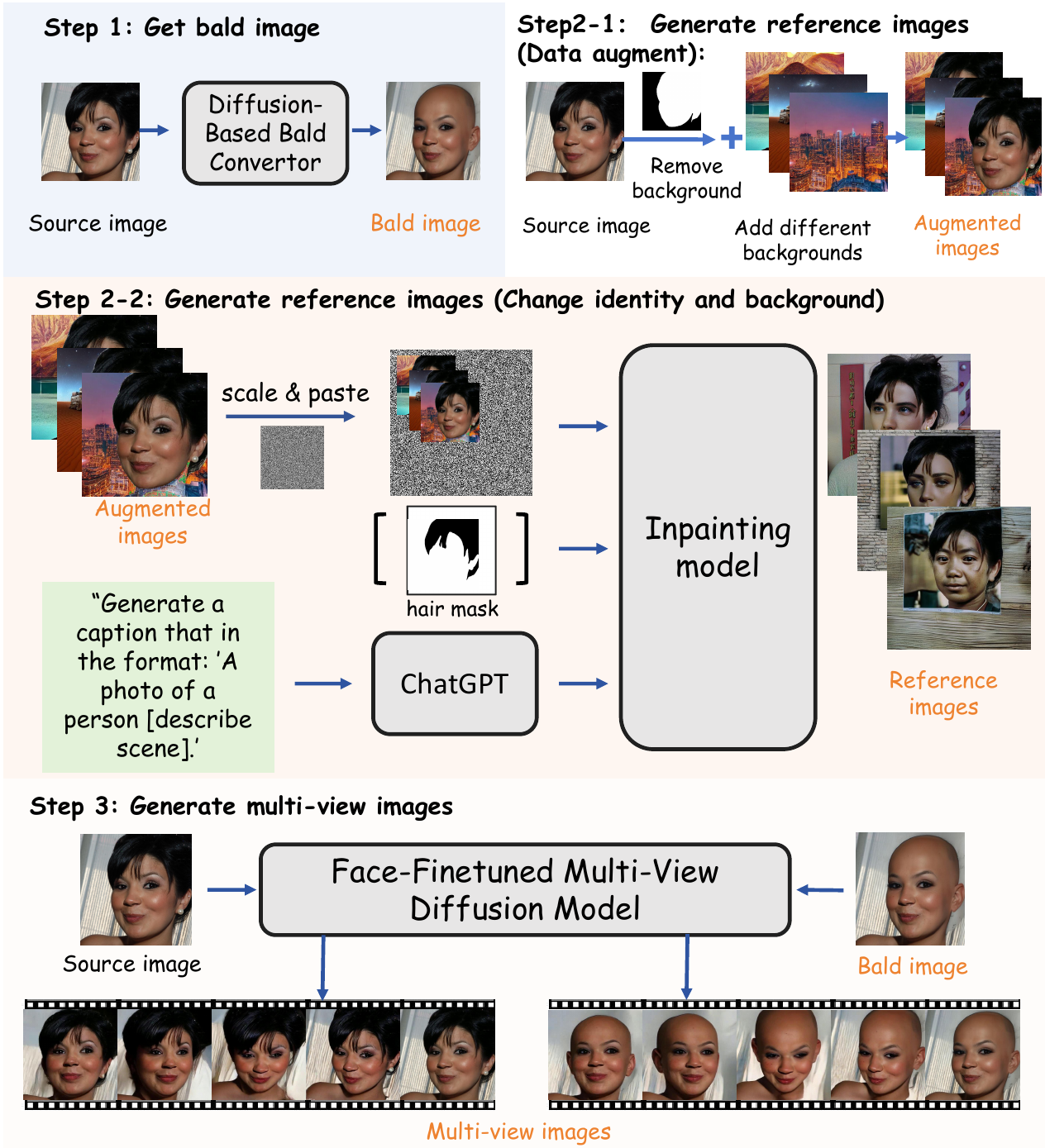}
    \vspace{-0.6cm}
    \caption{\textbf{Multi-View Training Data Generation Pipeline:} This pipeline aims to generate \{Multiple-View Original Image, Reference Image, Multiple-View Bald Image\} triplets for training. The pipeline includes three key steps. Step 1: use Diffusion-based Bald Convertor to convert original image to Bald Image; Step 2: First, augment the original with different background, then utilize Diffusion-Based Inpainting Model to change identity and background of the augmented images. Step 3: utilize Face-Finetuned Multiple-View Diffusion Model to convert both original and bald images into multiple-view images. }
    \vspace{-0.3cm}
    \label{fig:data}
\end{figure}

To generate effective reference images $I_{r}$, we ensure they exhibit diverse identities and backgrounds while maintaining the same hairstyle as $I_{s}$. This paired training data enhances the robustness of our hairstyle transfer model, allowing it to better handle the complex variations in reference hairstyles encountered in real-world scenarios. To modify the identity and background of $I_{s}$ while preserving the original hairstyle, we leverage hairstyle masks to isolate hair regions. Using a Stable Diffusion inpainting model guided by ChatGPT-generated prompts, we edit the non-hairstyle regions—such as facial features and backgrounds—of the original dataset (Fig.~\ref{fig:data}). Additionally, we introduce variations in scale to further diversify the reference images in terms of facial size and composition. This process produces a rich set of reference images $I_{r}$ for training, where hairstyles remain unchanged while identities, backgrounds, and spatial scales vary significantly. 

\begin{figure}[!t]
    \centering
    \includegraphics[width=1\linewidth]{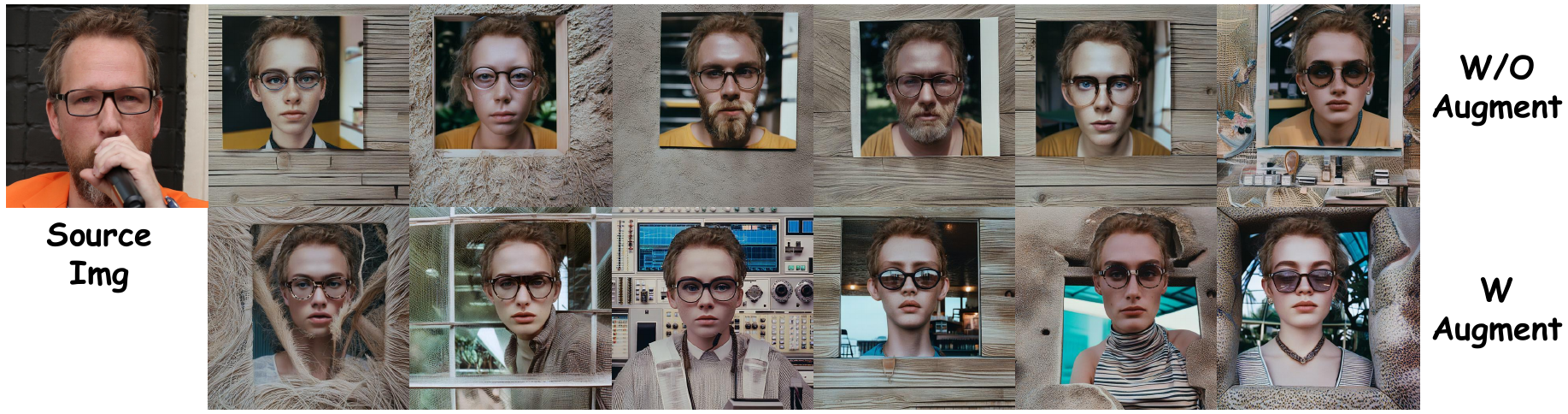}
    \vspace{-0.7cm}
    \caption{The reference images generated with and without data augmentation are shown in the figure. Without augmentation, the generated images often exhibit yellow shirts or backgrounds around the neck area, similar to the original image.}
    \vspace{-0.3cm}
    \label{fig:dataaug}
\end{figure}

As shown in Fig.~\ref{fig:dataaug}, we observed that the inpainting model often fails to completely exclude non-hairstyle-related information of $I_{s}$ from $I_{r}$. For the multi-view hairstyle transfer model, the bald image $I_{b}$ input during training often has a viewpoint inconsistent with that of $I_{r}$. This inconsistency may cause the hair encoder to inadvertently learn residual non-hairstyle information from $I_{r}$ that the inpainting model failed to remove, thereby compromising the hairstyle transfer quality.

To address this issue, we enhance the original pipeline by introducing a data augmentation strategy, as illustrated in Fig.~\ref{fig:data} (Step2-1). Specifically, we first generate 100 diverse background images with varying styles and scenery using ChatGPT combined with Stable Diffusion. Next, we extract the facial and hair regions of the source image using a mask and composite these regions with the generated backgrounds to create varied source images. Finally, we apply the inpainting model to these augmented source images to produce reference images. To ensure the hair encoder effectively learns hairstyle features, we generate 10 distinct reference images $\{I_r^i\}_{i=1}^{10}$ for each source image by pairing it with 10 different backgrounds. During training, a single $I_r$ is randomly selected from these 10 options for each of $I_{s}$.

\subsubsection{Face-Finetuned Multi-View Diffusion Model} Training the model requires generating multi-view image pairs ${(I_s^i, I_b^i)}_{i=1}^{K}$ for both the source and bald representations. However, due to the absence of publicly available multi-view video datasets for human faces, a common approach is to synthesize such data using 3D-aware GANs~\cite{Chan2022,xiang2023gramhd} combined with GAN inversion techniques~\cite{ko20233d}.While effective, GAN inversion often suffers from inefficient optimization and inherent reconstruction inaccuracies, which can compromise the quality and consistency of the resulting multi-view data.

As illustrated in Fig.~\ref{fig:data} (Step 3), we instead employ a pretrained SV3D model~\cite{voleti2024sv3d} to generate the corresponding multi-view images. While SV3D serves as a general-purpose solution for multi-view generation, it exhibits limitations in view reconstruction accuracy and controllability when applied to human faces. To address this, we introduce the Face-Finetuned Multi-View Diffusion Model, which fine-tunes the base model on facial data. Specifically, we first generate approximately 20,000 sets of multi-view face videos using a state-of-the-art 3D-aware GAN method~\cite{xiang2023gramhd}. These videos are then used to fine-tune the SV3D model. After fine-tuning, both the source image and the bald image are fed into SV3D to produce their respective multi-view images.

\subsection{Multi-View Hair Transfer Model with Multiple Stage Training Strategy}

The multi-view hair transfer model employs a multiple stage training strategy: (1) training a Pose-Controllable Latent IdentityNet with
multi-view data for view control; (2) training a Hair Extractor Module using a triplet dataset to transfer detailed reference hair onto a bald proxy image; and
(3) incorporating and training a Temporal Attention Module within the UNet blocks to ensure view-consistent outputs.

\subsubsection{Training Pose-Controllable Latent IdentityNet}

To enable multi-view generation, a key requirement is introducing pose controllability into the generative model. Inspired by recent multi-view diffusion approaches~\cite{voleti2024sv3d, liu2023zero}, we propose a pose-controllable IdentityNet integrated into the denoising UNet, as shown in Fig.~\ref{fig:method} (Left). This module encodes polar and azimuth angles ($p$,$a$) into the sinusoidal embedding, which is then added to the time embedding to attain the new embedding $e_{f}$:

\begin{equation}
    e_{f} = \mathbf{E}(t) + C[ \mathbf{E}(\epsilon), \mathbf{E}(p),\mathbf{E}(a)]
\end{equation}
where $\mathbf{E}$ is a sinusoidal embedding, $C$ denotes the Concatenation operation and $\epsilon$ is a noise augmentation of camera pose. 

Initially, we adopt the ControlNet architecture as our IdentityNet. While ControlNet effectively preserves structural consistency with the source image, our experiments reveal that it struggles with color consistency. As shown in Fig.~\ref{fig:ablation}, accumulated color deviations across multiple generation steps result in noticeable shifts in final image colors.

To address this, we propose a Latent IdentityNet, which extends IdentityNet by introducing a VAE encoder to project the input image into latent space. During training, we randomly select one frame from the multi-view bald dataset as the input and another as the ground truth. The corresponding polar and azimuth angles for both frames are provided to the model as pose conditions. Importantly, we freeze all other network components and train only the IdentityNet, ensuring targeted learning of pose-conditioned identity preservation.

\subsection{Training hair extractor}

We first train the model to learn pose alignment using bald multi-view images, which eliminates interference from hair details and promotes more consistent head shape modeling. Once reliable pose control is achieved, we proceed to train the hairstyle transfer module, which synthesizes hair conditioned on the aligned head representations. Inspired by recent advancements in reference image-guided generation~\cite{stablegarment,zhang2024flashface}, our multi-view hair transfer framework incorporates a trainable U-Net, initialized from a pre-trained diffusion model, which we designate as the Hair Extractor. As shown in Fig.~\ref{fig:method} (Middle), $I_{r}$ is encoded through the Hair Extractor, and the features from the self-attention layers in each block are extracted to capture fine-grained hair details. These detailed hair features are then integrated into the diffusion U-Net via newly introduced hair cross-attention layers. Within each block of the U-Net, we preserve the original self-attention layers while augmenting them with hair cross-attention layers. The extracted hair features are utilized as the key ($K$) and value ($V$) inputs in the hair cross-attention layers, while both the self-attention and hair cross-attention layers share the same query ($Q$) feature. In this phase, we freeze the parameters of the pose-controllable IdentityNet and exclusively train the parameters of Hair Extractor. Specifically, we randomly select one frame $I_{b}^{i}$ from multiple-view bald images, along with the $I_{r}$ of the hairstyle generated for the corresponding identity. Additionally, we randomly sample polar and azimuth angles ($p_{i}$,$a_{i}$), and use the corresponding source image as the ground truth $I_{g}$.


\begin{figure*}[!t]
    \centering
    \includegraphics[width=1.00\linewidth]{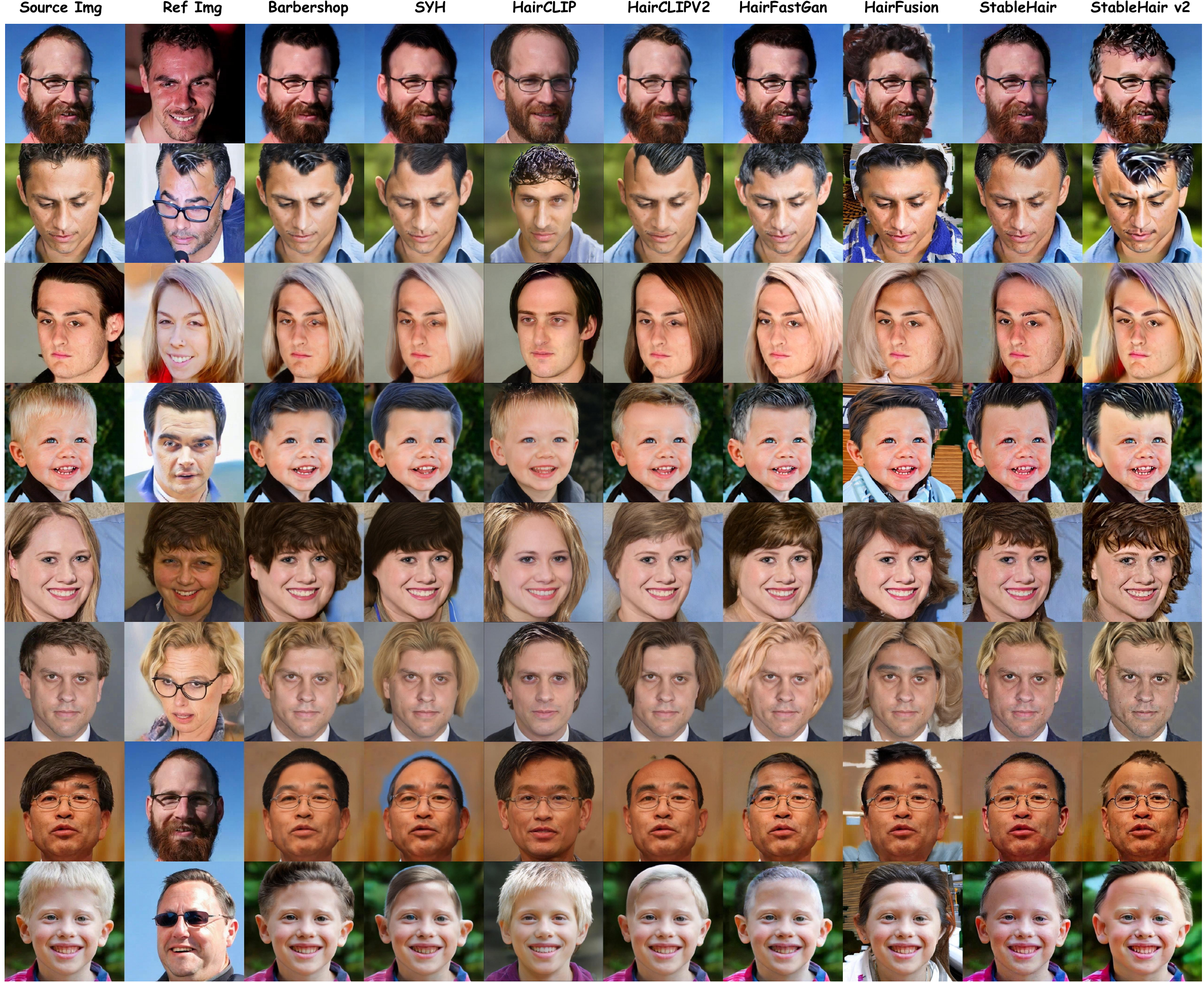}
    \vspace{-0.5cm}
    \caption{Qualitative comparison of different methods. Compared to other approaches, our method achieves more refined and stable hairstyle transfer without the need for precise facial alignment or explicit masks for supervision.}
    \vspace{-0.3cm}
    \label{fig:com}
\end{figure*}

\begin{figure*}[!t]
    \centering
    \includegraphics[width=1.00\linewidth]{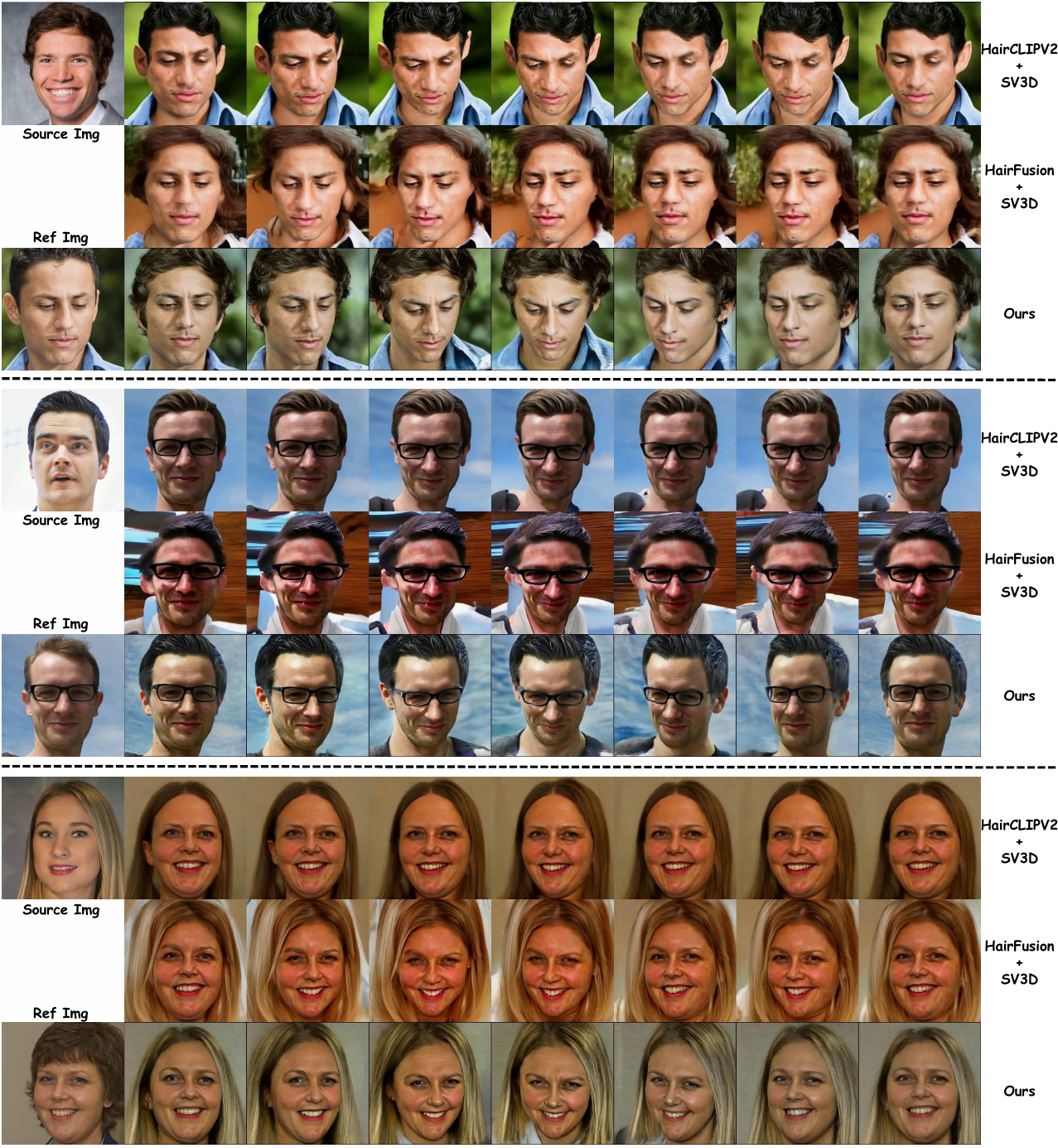}
    \vspace{-0.5cm}
    \caption{Qualitative comparison of different multi-view hair transfer pipeline. Compared to other pipeline, our method achieves more refined and stable hairstyle transfer without the need for precise facial alignment or explicit masks for supervision.}
    \vspace{-0.3cm}
    \label{fig:com2}
\end{figure*}

\begin{figure*}[!t]
    \centering
    \includegraphics[width=1.00\linewidth]{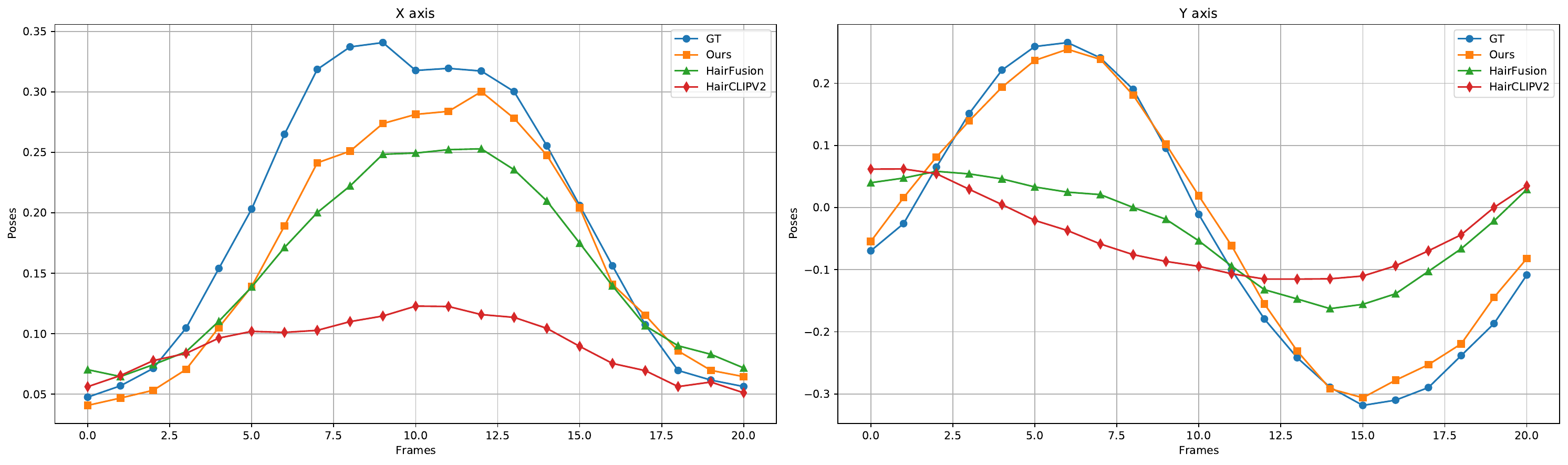}
    \vspace{-0.25cm}
    \caption{The head motion along the x-axis and y-axis of the three-dimensional coordinate system of three methods and ground truth.}
    \vspace{-0.3cm}
    \label{fig:line}
\end{figure*}


\begin{figure*}[h]
    \centering
 \includegraphics[width=1.0\linewidth]{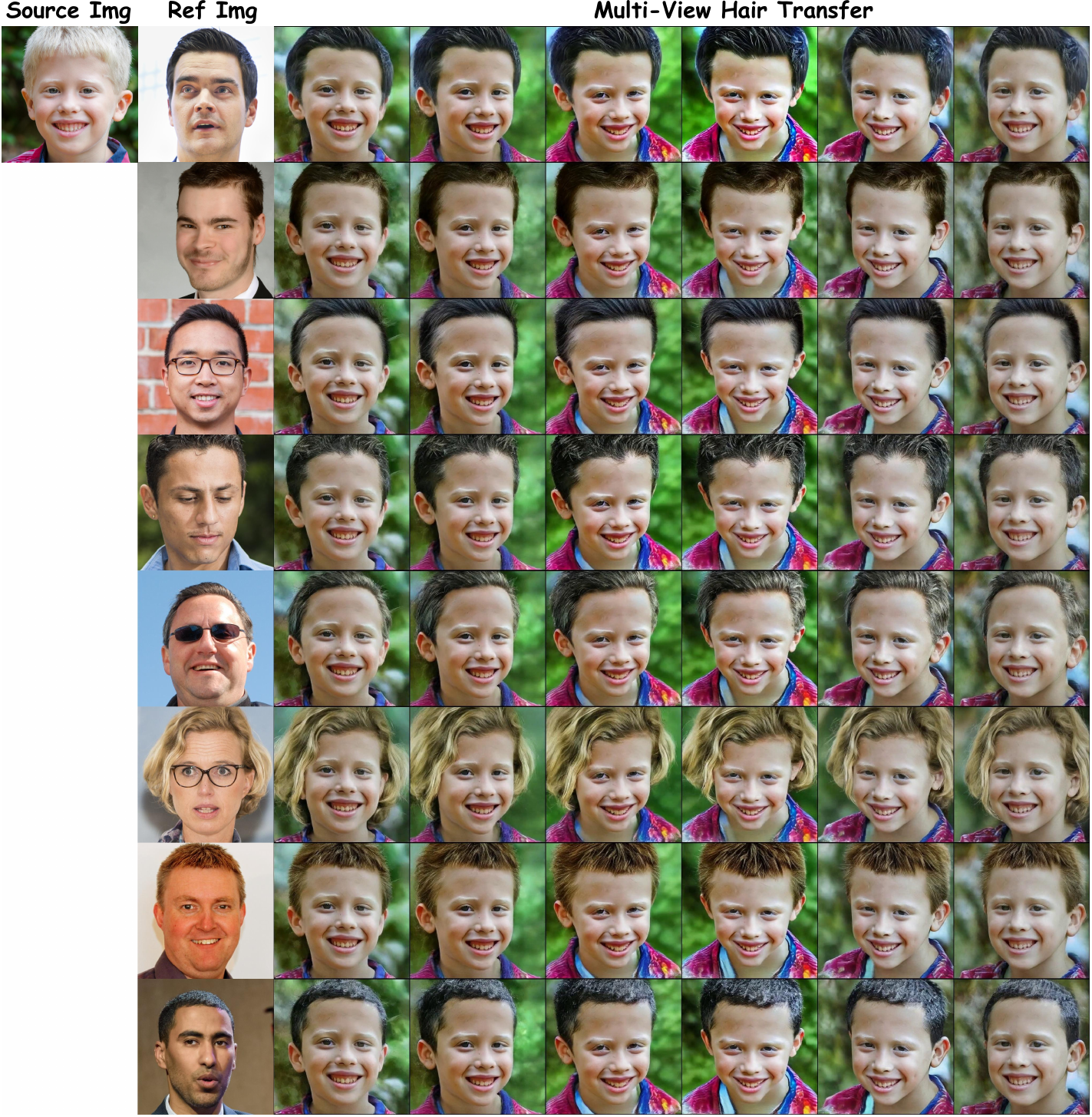}
    \caption{Visualization of mult-view hair transfer. Our methods can not only transfer ref hairstyle to the original img, but also generate corresponding multi-view images. }
    \vspace{-0.6cm}
    \label{fig:show}
\end{figure*}

\subsection{Training Temporal Attention}

After the first two training phases, the model is capable of generating hairstyle transfer results for specific viewpoints. However, when generating multiple consecutive frames, temporal consistency across frames remains suboptimal. To address this, we draw inspiration from AnimateDiff~\cite{guo2023animatediff} and introduce a temporal attention mechanism. Specifically, we insert a temporal self-attention block after each block in the U-Net of Stable Diffusion, as illustrated in Fig.~\ref{fig:method} (Right).

Unlike vanilla Stable Diffusion, our modified version incorporates an additional sequence dimension in the input, allowing the model to process multiple frames simultaneously. While the original attention layers in Stable Diffusion perform self-attention or cross-attention along spatial dimensions (height and width), the temporal attention module operates along the sequence dimension, capturing inter-frame dependencies.

During this third training phase, we freeze the parameters of both the pose-controllable IdentityNet and the Hair Extractor, and train only the temporal attention modules. For each training iteration, we randomly sample $k$ consecutive hair frames from $\{I_s^i\}_{i=1}^{K}$ as ground truth, along with a random $I_{b}$, the corresponding pose $(p_{i}, a_{i})_{i=1}^{k}$ for sampling frames, and a reference hairstyle image $I_{r}$ as input. This design encourages the model to learn consistent hairstyle generation across time.

\subsection{Model Training}


The loss functions for both phases can be mathematically represented as follows:
\begin{equation}
    L(\bm{\theta}) := \mathbb{E}_{\mathbf{x_0}, t, \bm{\epsilon}}\left[\left\|\bm{\epsilon} - \bm{\epsilon_\theta}\left(\mathbf{z_t}, t, \mathbf{c_s}, \mathbf{c_r}, \mathbf{c_v}\right)\right\|_2^2\right],
\end{equation}
where $\mathbf{z_t}$ is a noisy latent obtained by adding Gaussian noise $\bm{\epsilon} \sim \mathcal{N}(\mathbf{0}, \mathbf{1})$ to the clean image latent. The network $\bm{\epsilon_\theta}(\cdot)$ is trained to predict the added noise. 

$\mathbf{c_s}$, $\mathbf{c_r}$, and $\mathbf{c_v}$ represent the source condition (either the original image or the bald proxy image), the reference condition, and the viewpoint condition (encoded as polar and azimuth angles), respectively. 

During bald converter training, $\mathbf{c_r}$ is set to None, meaning only the source and viewpoint conditions are used. This unified objective enables both hairstyle removal and pose-controllable generation to be learned within the same noise prediction framework.

\section{Experiments}
\subsection{Implementation Details} 

We employed Stable Diffusion V1.5 as the pre-trained diffusion model. During the data preparation phase, we first applied a bald converter—pre-trained on the Non-Hair FFHQ dataset~\cite{Wu_2022_CVPR}—to transform the original FFHQ images into bald versions. To obtain multi-view data, we used a SV3D model fine-tuned on a 21-view facial dataset to generate 21 viewpoints for both the original and the bald images. However, some FFHQ images have extreme or uncommon viewpoints that deviate from the distribution of views in the fine-tuning dataset, leading to low-quality multi-view synthesis. To address this issue, we filtered out these outliers, resulting in approximately 20,000 high-quality multi-view image pairs. With the curated dataset, we first trained a pose-controllable latent IdentityNet using the multi-view images. This training was conducted on a single H800 GPU, using a batch size of 16 and a learning rate of 1e-5, for a total of 8,000 steps. Next, we trained a Hair Extractor network using both the reference images and the multi-view samples. This stage was performed on 8 H800 GPUs, with a batch size of 16, learning rate of 1e-5, and lasted for 100,000 steps. Finally, we trained a temporal attention module to enhance temporal consistency, using reference images and multi-view image pairs. This stage was also conducted on 1 H800 GPU, with a batch size of 1, image sequence length of 12, and a learning rate of 1e-5, over 30,000 optimization steps. During inference, we adopted the DDIM sampler with 30 sampling steps, and set the classifier-free guidance scale to 1.5.

\setlength{\tabcolsep}{0.9mm}{
\begin{table}[!t]
\centering
\caption{\textbf{Quantitative comparison} of different methods for single-view hair transfer. Metrics that are bold and underlined represent methods that rank 1st and 2nd, respectively.}
\vspace{-0.2cm}
\label{tab1}
\begin{tabular}{*{2}c|ccccc}
\toprule
& Method & CLIP-I $\uparrow$ & FID $\downarrow$ & PSNR $\uparrow$ & SSIM $\uparrow$ & IDS $\uparrow$\\
\midrule
&Barbershop (TOG 2021) &0.427 &46.103 &30.274 &0.607 &0.740\\
&SYH (ECCV 2022) & 0.419 &40.513 &30.478 &0.635 &0.707\\
&HairCLIP (CVPR 2022) &0.383  &43.377 &28.759 &0.605 &0.691\\
&HairCLIPV2 (ICCV 2023) &0.407 &37.256 &30.482 &0.632&0.759\\
&HairFastGAN (NeurIPS 2024) & 0.416&36.403  &30.098 &0.654 &0.758\\
&HairFusion (AAAI 2025) &0.416 &36.302 &\underline{30.646} &\underline{0.664}&\underline{0.762}\\
\cdashline{1-7}
&StableHair (AAAI 2025) &\underline{0.429} &\underline{36.128} &30.564 &\underline{0.670} &\underline{0.776} \\
&StableHair v2 &\textbf{0.431} &\textbf{35.125} &\textbf{30.671} &\textbf{0.673} &\textbf{0.778} \\
\bottomrule
\end{tabular}
\label{table:1}
\vspace{-0.3cm}
\end{table}
}

\subsection{Evaluation Metrics}
Given a hairstyle reference image, the purpose of hair transfer is to apply the corresponding hairstyle and hair color attributes to the input image. We compare our method with current state-of-the-art hair transfer methods:
\textit{Barbershop} \cite{zhu2021barbershop}, \textit{SYH} \cite{kim2022styleSYH}, \textit{HairFastGAN} \cite{nikolaev2024hairfastgan}, \textit{HairCLIP} \cite{wei2022hairclip}, \textit{HairCLIPV2} \cite{wei2023hairclipv2}, \textit{HairFusion} \cite{chung2025hairfusion} and StableHair~\cite{zhang2025stable}. All comparison algorithms use the default parameters from their official implementations.

To provide a comprehensive and objective evaluation of each algorithm’s performance in various aspects of hairstyle transfer, we compute the Fréchet Inception Distance (FID)\cite{fid} between the source image and the generated image to assess overall visual quality. Since hairstyle transfer is expected to preserve the identity and background of the source image, we evaluate the identity and background consistency using SSIM\cite{wang2004imageSSIM} and PSNR. Specifically, SSIM and PSNR are computed over the intersected non-hair regions before and after editing to focus on areas that should remain unchanged. To further assess identity preservation, we employ InsightFace\cite{arcface} to compute the Identity Similarity (IDS) between the original source image and the generated target image. Additionally, to measure the effectiveness of hairstyle transfer itself, we use the CLIP-I\cite{clip} metric, which calculates the cosine similarity between the image embeddings of the reference hairstyle image and the generated result.

We conducted two sets of experiments: one for single-view hairstyle transfer and another for multi-view hairstyle transfer. For the single-view hairstyle transfer, we compared our method with all above-mentioned approaches. Since there are currently no dedicated methods for multi-view hairstyle transfer, we selected the two most robust hairstyle transfer methods (HairCLIPV2 and HairFusion) based on their overall performance. These methods were first used to generate single-image hairstyle transfer results, which were then fed into a fine-tuned SV3D model to produce the corresponding multi-view videos.

During the experiments, we randomly selected \textbf{50} distinct samples to constitute the test set, performed cross-sample hairstyle transfer among them, and evaluated the corresponding metrics.


\subsection{Experiment Results}
\subsubsection{Qualitative Comparison.}
The qualitative comparison experiments of single-view hair transfer across a variety of hairstyles are presented in Fig.\ref{fig:com}. Overall, our method significantly outperforms other approaches in terms of the fidelity and coherence of hairstyle transfer. Among the compared methods, Barbershop\cite{zhu2021barbershop} often produces unstable results for complex hairstyles, as seen in the second and fifth rows. SYH~\cite{kim2022styleSYH} and HairFastGAN~\cite{nikolaev2024hairfastgan} tend to generate relatively coarse outputs, frequently missing fine-grained hair texture and color details. HairCLIP~\cite{wei2022hairclip} and HairCLIPV2~\cite{wei2023hairclipv2} show the weakest transfer capability, struggling to accurately replicate the reference hairstyle. Hairfusion~\cite{chung2025hairfusion} performs poorly in preserving the background and maintaining source identity consistency. In contrast, our StableHair v2 consistently demonstrates robust performance across a wide range of hairstyle types and colors, similar to the original StableHair.

\setlength{\tabcolsep}{0.9mm}{
\begin{table}[!t]
\centering
\caption{\textbf{Quantitative comparison} of different methods for multi-view hair transfer. Metrics that are bold and underlined represent methods that rank 1st and 2nd, respectively.}
\vspace{-0.2cm}
\label{tab1}
\begin{tabular}{*{2}c|ccccc}
\toprule
& Method & CLIP-I $\uparrow$ & FID $\downarrow$ & PSNR $\uparrow$ & SSIM $\uparrow$ & IDS $\uparrow$\\
\midrule
&HairCLIPV2 + SV3D &0.397 &29.652 &25.582 &0.432&0.632\\
&HairFusion + SV3D &\underline{0.406} &\underline{31.203} &\textbf{27.846} &\textbf{0.564}&\underline{0.659}\\
\cdashline{1-7}
&Ours &\textbf{0.411} &\textbf{32.170} &\underline{26.347} &\underline{0.490} &\textbf{0.683} \\
\bottomrule
\end{tabular}
\label{table:2}
\vspace{-0.3cm}
\end{table}
}

The comparison results of multi-view hairstyle transfer are shown in Fig.~\ref{fig:com2}. It can be observed that our method achieves superior hairstyle accuracy, whereas the hairstyle transfer results from HairCLIPV2 and HairFusion are less precise, leading to inaccuracies in the hairstyles of the multi-view images generated by SV3D. To demonstrate the accuracy of different methods in controlling head motion, we computed the 3DMM parameters for each frame of the videos generated by the three methods as well as the ground truth using Deep3DFaceReconstruction~\cite{deng2019accurate}. We then visualized the x-axis and y-axis 3DMM pose parameters for each frame, as shown in the Fig.~\ref{fig:line}. We observe that our method's head motion aligns most closely with the ground truth. HairCLIPV2 and HairFusion generate multi-view results by inputting 2D images obtained through hairstyle transfer into SV3D. Although the input angles in SV3D are consistent with those in the ground truth, the results from these two methods still exhibit significant discrepancies compared to the ground truth. This is primarily due to the gap between the 2D hairstyle transfer results and the facial dataset used during the fine-tuning of SV3D, which introduces certain errors in the generated multi-view outcomes.



Fig.~\ref{fig:show} presents additional results of multi-view hairstyle transfer. In this figure, we demonstrate the transfer of different hairstyles to the same source image. As can be observed, our method effectively adapts to various hairstyles in the multi-view hairstyle transfer task. More results can be found at our demo video.




\subsubsection{Quantitative Comparison.}
For the quantitative experiments, we utilized a dataset generated by our training data pipeline. Specifically, we employed 2,500 sets of multi-view images, from each of which we randomly selected one frame as the source image and randomly chose one frame from another 2,500 sets of multi-view images as the reference image for hairstyle transfer. For multi-view hairstyle transfer experiment, our method takes the source image and the corresponding reference image as input to generate a multi-view video of the source image’s identity with the transferred hairstyle. In contrast, the other two comparison pipelines first perform hairstyle transfer on the source image and then use SV3D to generate a multi-view video of the transferred image.

Table~\ref{table:1} presents the quantitative comparison of various methods on the single-view hairstyle transfer task. Overall, StableHair v2 consistently outperforms existing approaches across all evaluation metrics. In particular, it achieves the highest PSNR and FID scores, indicating superior fidelity and perceptual quality. Moreover, the top CLIP-I score demonstrates that our method effectively aligns the transferred hairstyle with the reference image, reflecting strong style consistency. The highest IDS score further highlights our method’s ability to preserve identity, achieving the best identity retention among all compared methods.


\setlength{\tabcolsep}{1mm}{
\begin{table}[t]
\centering
\caption{\textbf{Quantitative ablation} of different design options. (`pixel' refers to the use of the original ControlNet, while 'latent' denotes the use of the Latent ControlNet.)}
\vspace{-0.3cm}
\label{tab:ab}
\begin{tabular}{*{2}c|ccccc}
\toprule
& Method & CLIP-I $\uparrow$ & FID $\downarrow$ & PSNR $\uparrow$ & SSIM$ \uparrow$ & IDS $\uparrow$\\
\midrule
&Ours (latent) &\textbf{0.430} &\textbf{35.128} &\textbf{30.567} &\textbf{0.670} &\textbf{0.773} \\
&Ours (pixel)&0.403 &41.134 &29.014  &0.629 &0.733 \\
\bottomrule
\end{tabular}
\vspace{-0.3cm}
\end{table}
}

Table~\ref{table:2} shows our quantitative evaluation across different methods on multi-view hairstyle transfer experiment. Our method not only addresses the limitations of existing approaches but also demonstrates superior performance in terms of FID and IDS. However, it falls short in SSIM and PSNR metrics compared to the single-view hairstyle transfer experiment. This discrepancy is primarily attributed to the suboptimal background reconstruction in our multi-view hairstyle transfer process.

\subsubsection{User Study.}
Given the subjective nature of the hairstyle transfer task, we conducted a comprehensive user study with 30 volunteers. Specifically, we randomly sampled 20 data pairs from our quantitative experiments and selected 10 popular hairstyles from social media as reference styles. For each reference, a corresponding source image was randomly selected from the FFHQ dataset, resulting in an additional 10 data pairs. In total, we created 30 triplets, each consisting of an original image, a reference image, and both single-view and multi-view transfer results. Following the protocol of prior work \cite{wei2022hairclip}, the outputs from different methods were presented in a randomized order to avoid bias. For each single-view transfer sample, participants were asked to evaluate and select the best result based on three criteria: transfer accuracy, smoothness of head motion, and visual naturalness. Accuracy denotes the accuracy for hair transfer, Preservation indicates the ability to preserve irrelevant regions and Naturalness denotes the visual realism of the generated image. 

As shown in Table~\ref{tab2}, the results of the single-view hairstyle transfer indicate that our method outperforms existing approaches across all three evaluation criteria. Similarly, the results of the multi-view hairstyle transfer, presented in Table~\ref{tab3}, further confirm that our method consistently surpasses the baselines in terms of transfer accuracy, motion smoothness, and overall visual realism.

\setlength{\tabcolsep}{0.6mm}{
\begin{table}[t]
\centering
\caption{\textbf{User study} on single view hair transfer.}
\vspace{-0.3cm}
\begin{tabular}{*{2}c|ccccccc}
\toprule
&\multirow{2}*{Metrics}  &Barber  &Hair  &\multirow{2}*{SYH}  &Hair  &Hair &Stable &Stable \\
& &shop &FastGAN & &CLIP &CLIPV2 & HairV1 & HairV2 \\
\midrule
&Accuracy(\%) &13.6  &15.1  &13.2  &5.1  &10.2  &20.9&\textbf{21.9}\\
&Preservation(\%) &11.1  &11.3  &14.0  &7.4  &15.8  &20.1& \textbf{20.3}\\
&Naturalness(\%) &11.4  &14.2  &13.9  &11.2  &14.0  &17.6 &\textbf{17.7}\\
\bottomrule
\end{tabular}
\vspace{-0.4cm}
\label{tab2}
\end{table}
}

\setlength{\tabcolsep}{0.6mm}{
\begin{table}[t]
\centering
\caption{\textbf{User study} on multi-view hair transfer.}
\vspace{-0.3cm}
\begin{tabular}{*{2}c|ccc}
\toprule
&Metrics     & Hair CLIP V2+SV3D  & HairFusion+SV3D & Ours \\
\midrule
&Accuracy(\%)       &26.1  &28.3  &\textbf{46.6}\\
&Smoothness(\%)      &26.8  &28.9  &\textbf{44.3}\\
&Naturalness(\%)    &25.3  &27.7  &\textbf{47.0} \\
\bottomrule
\end{tabular}
\vspace{-0.4cm}
\label{tab3}
\end{table}
}

\begin{figure}[!t]
    \centering
    \includegraphics[width=1\linewidth]{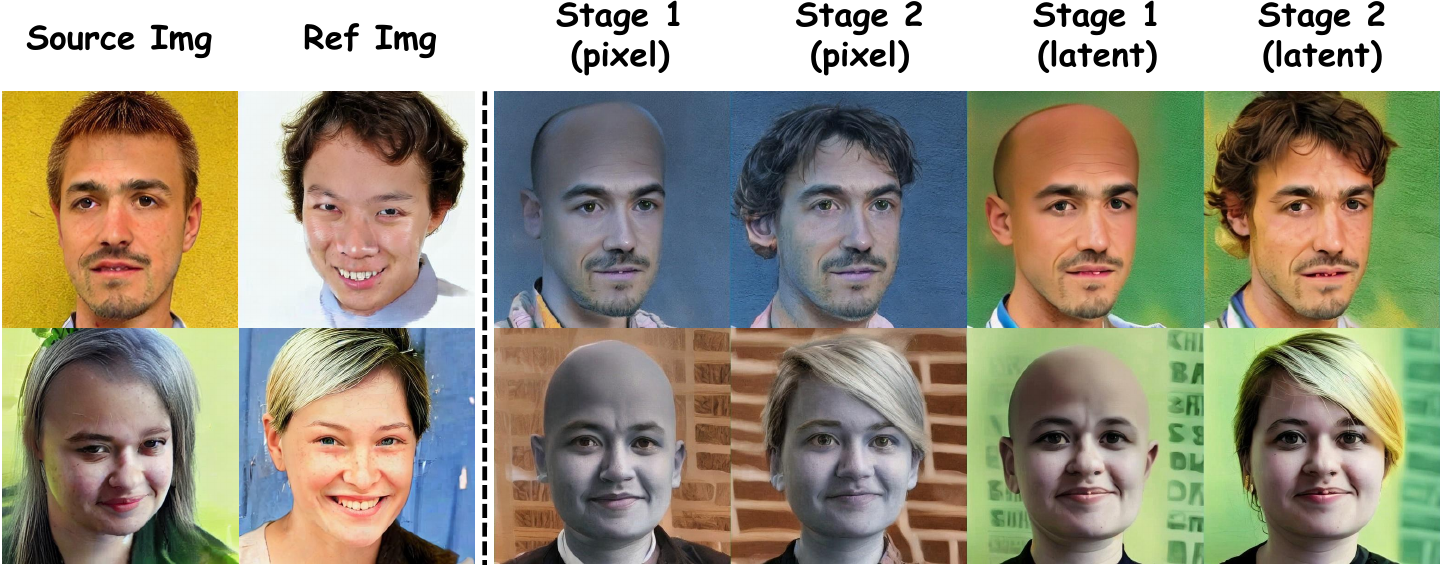}
    \vspace{-0.7cm}
    \caption{Visual results of the ablation study comparing different architectural variants. The proposed Latent ControlNet effectively preserves color consistency, while the original pixel-space ControlNet introduces noticeable color shifts across stages. Stage 1 and Stage 2 correspond to the two-step process illustrated in Fig.~\ref{fig:method}.}
    \vspace{-0.3cm}
    \label{fig:ablation}
\end{figure}

\subsubsection{Ablation Study.} 
To comprehensively assess the contribution of each module in our framework, we conduct a systematic ablation study encompassing the Pose-Controllable Latent ControlNet, Data-Augmented Inpainting Model, Stage-Wise Training and Temporal Attention Module. As shown in Fig.~\ref{fig:ablation}, models trained with pixel-space ControlNet often suffer from color inconsistencies between source and target images (3rd and 4th columns), due to direct pixel-conditioned guidance. In contrast, our proposed Pose-Controllable Latent ControlNet transfers the hair editing process into latent space, effectively mitigating such inconsistencies and substantially improving content fidelity (5th and 6th columns). Quantitative results in Table~\ref{tab:ab} further confirm the advantage: Latent ControlNet consistently achieves superior performance across all metrics, demonstrating stronger transfer ability while preserving identity details more faithfully.

\begin{figure}[!t]
    \centering
    \includegraphics[width=1\linewidth]{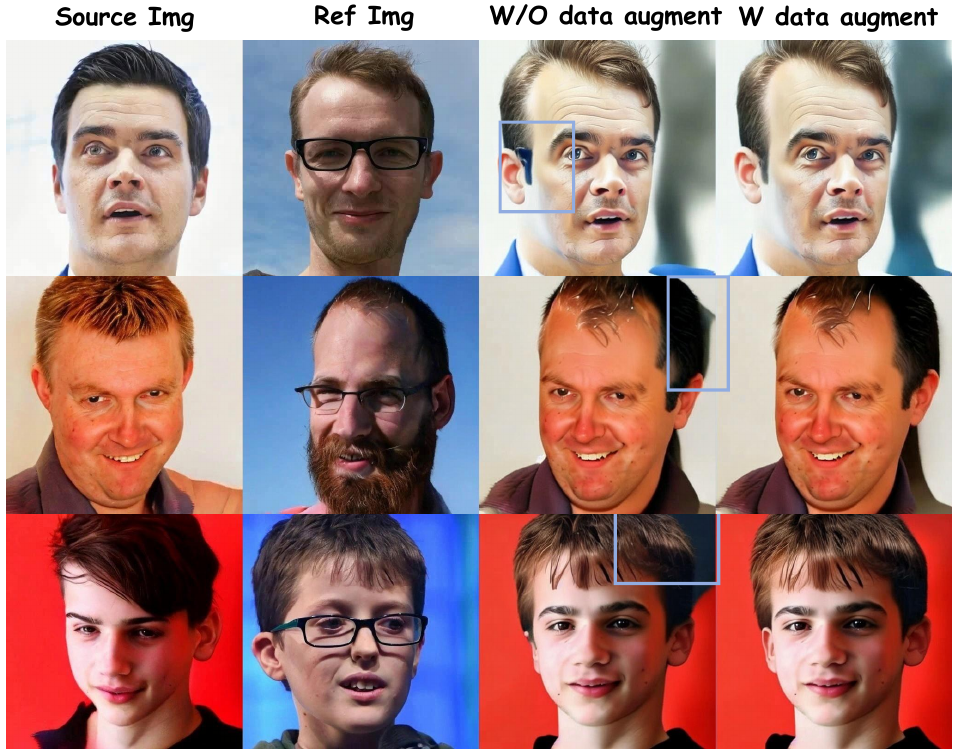}
    \vspace{-0.6cm}
    \caption{\textbf{Data Augmentation Ablation} results of different training data. When the model is trained with augmented data, the boundaries of the generated hairstyles tend to be more distinct compared to those trained without data augmentation.}
    \label{fig:ab_v}
    \vspace{-0.3cm}
\end{figure}

To further evaluate the effectiveness of data augmentation in our Data-Augmented Inpainting Model, we trained a variant using reference images without augmentation. As shown in Fig.~\ref{fig:ab_v}, incorporating data augmentation significantly improves the accuracy of hairstyle transfer. As indicated by the blue bounding boxes, the model trained without augmented references tends to generate artifacts, resulting in hairstyles that blend indistinctly with the background. In contrast, the model trained with augmented references produces cleaner boundaries and sharper hairstyle structures.

\begin{figure}[!t]
    \centering
    \includegraphics[width=1\linewidth]{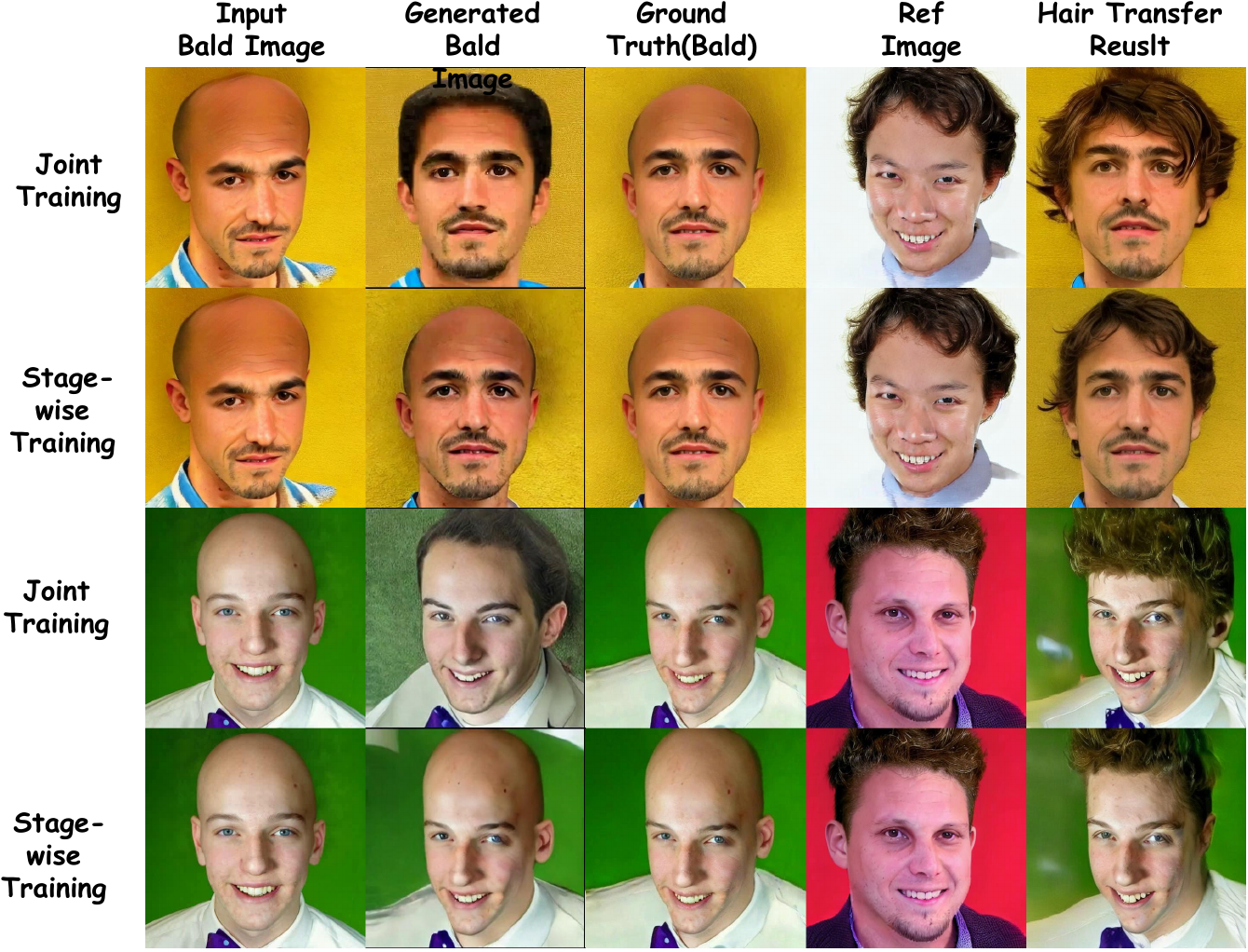}
    \vspace{-0.6cm}
    \caption{ Ablation comparison between joint and two-stage training strategies. The two-stage training first learns pose control independently, followed by hairstyle transfer, which leads to improved bald head translation and ultimately results in more accurate and consistent hairstyle transfer.}
    \label{fig:ab_t}
    \vspace{-0.3cm}
\end{figure}

We compare the proposed stage-wise training strategy with joint training of Stage 1 (pose control) and Stage 2 (hairstyle transfer). As shown in Fig.~\ref{fig:ab_t}, the joint training baseline exhibits suboptimal pose transfer for bald images (1st row, 2nd column), which results in inconsistent head shapes in the final hair transfer output (1st row, 5th column). In contrast, the stage-wise training first learns accurate pose control using bald multi-view inputs, followed by training the hairstyle transfer module. This decoupled learning process leads to improved bald head alignment (2nd row, 2nd column), which in turn enables more accurate and consistent hairstyle generation across views (2nd row, 5th column).

\begin{figure}[!t]
    \centering
    \includegraphics[width=1\linewidth]{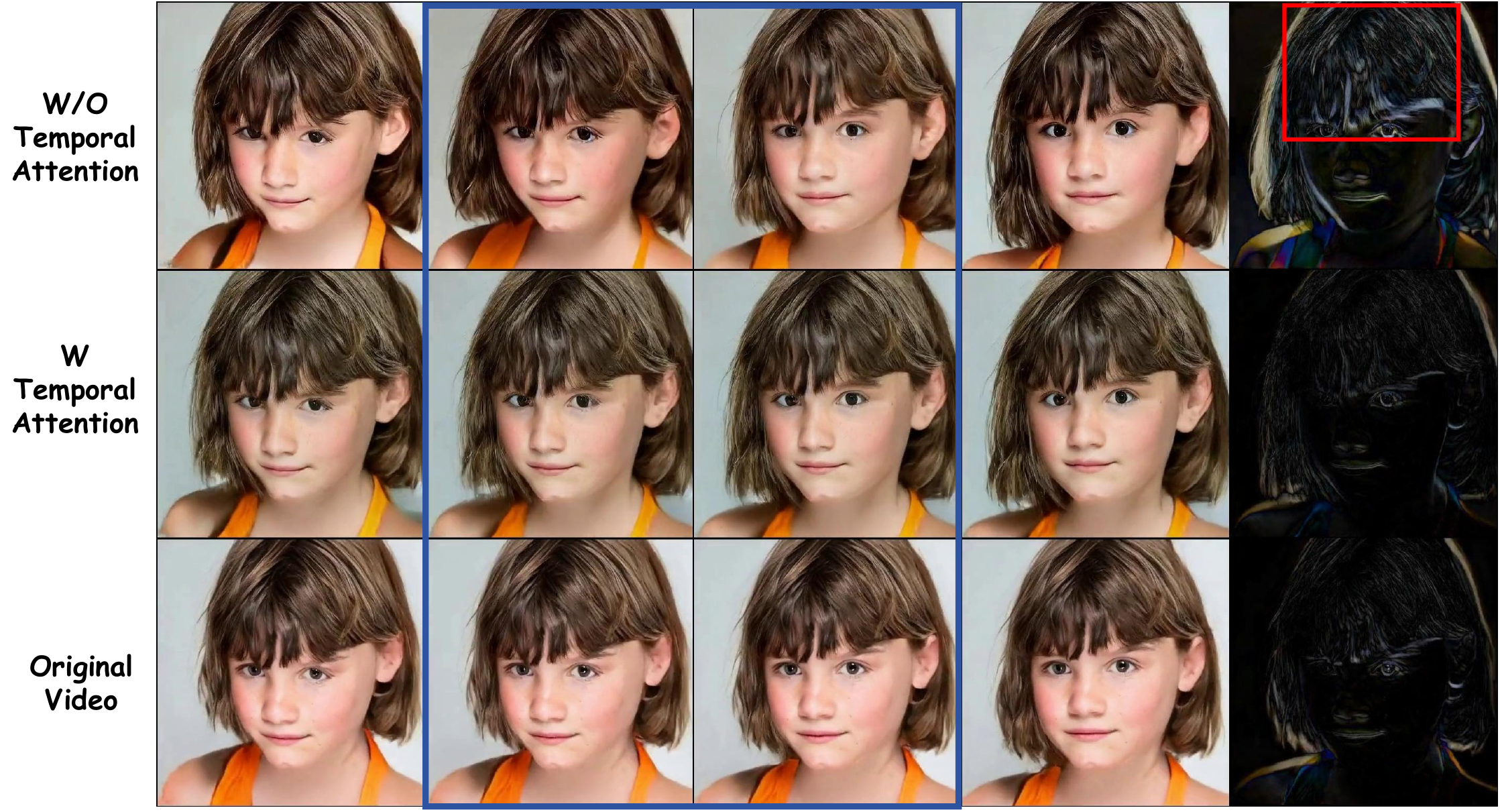}
    \vspace{-0.6cm}
    \caption{\textbf{Visual Ablation} results of temporal attention. The first four columns represent consecutive frames from the video, while the fourth column displays the heatmap of frame-wise changes. The heatmap for the video without temporal attention shows regions with higher brightness, corresponding to less smooth transitions in those areas.}
    \label{fig:ab_v2}
    \vspace{-0.3cm}
\end{figure}

We further conducted an ablation study on the Temporal Attention Module, with results presented in Fig.~\ref{fig:ab_v2}. The first four columns depict the input and three consecutive output frames, while the fifth column shows the heatmap of frame-wise differences. In the heatmaps, darker regions indicate minimal pixel variation, whereas brighter regions correspond to larger temporal changes, with brightness intensity reflecting the degree of variation. As observed, both the ground truth video and the model equipped with temporal attention layers exhibit predominantly dark regions in the hair areas, indicating smooth and stable transitions across frames. In contrast, the model without temporal attention displays brighter regions in these areas, suggesting notable fluctuations and reduced temporal consistency. These results underscore the effectiveness of our temporal attention mechanism in preserving motion coherence and hairstyle stability across frames. For additional visual comparisons, please refer to the supplementary material.

In addition, we conducted quantitative evaluations to assess the effectiveness of the Temporal Attention Module. Specifically, we applied GAN inversion using GRAM-HD to the first frame of videos generated by models with (W) and without (W/O) temporal attention, and synthesized corresponding videos. We then computed SSIM, PSNR, and LPIPS metrics between these generated videos and the original outputs, on a frame-by-frame basis. As shown in Table~\ref{tab_ablation}, the model equipped with temporal attention consistently outperforms its counterpart across all metrics, indicating improved temporal consistency and visual quality. These results demonstrate that temporal attention significantly enhances the smoothness of the generated sequences, bringing them closer in quality to those produced by dedicated multi-view video generation models. See the supplements for additional ablation studies.



\setlength{\tabcolsep}{0.8mm}{
\begin{table}[!t]
\centering
\caption{\textbf{ablation study} on temporal attention layers.}
\vspace{-0.3cm}
\begin{tabular}{*{2}c|ccc}
\toprule
&Metrics     & PSNR $\uparrow$  & SSIM $\uparrow$ & LPIPS $\downarrow$ \\
\midrule
&W Temporal Attention       & \textbf{17.945}  & \textbf{0.575}  & \textbf{0.163} \\
&W/O Temporal Attention      & 17.800  & 0.487  & 0.194\\
\bottomrule
\end{tabular}
\vspace{-0.4cm}
\label{tab_ablation}
\end{table}
}

\section{Limitations and Future Work}
While our method achieves high-quality multi-view hairstyle transfer, it still presents several limitations. One notable issue is suboptimal background reconstruction, particularly in terms of inter-frame coherence. This limitation primarily stems from the model's limited access to consistent background information across different viewpoints, leading to artifacts and inconsistencies between frames. Although the temporal attention mechanism improves temporal continuity, it remains insufficient for accurately restoring complex background details. To overcome this, future work could explore explicitly encoding background features—either through background-aware modules or viewpoint-invariant representations—to enable better background consistency and control. Another limitation lies in the restricted range of training views, which currently focus on front-facing perspectives. To support more comprehensive full-view generation, we plan to extend the training dataset to include a wider range of viewpoints, thereby enabling more robust and accurate synthesis from arbitrary angles.

\section{Conclusions}
In this paper, we propose Stable-Hair v2, the first diffusion-based framework for multi-view hairstyle transfer. Our method represents a substantial advancement in the field, enabling stable and fine-grained hairstyle transfer across multiple views in real-world scenarios—capabilities that were previously difficult to achieve. Stable-Hair v2 comprises a multi-view training data generation pipeline built on a three-step procedure, alongside a multi-stage training strategy designed for high-fidelity multi-view hair transfer. Extensive qualitative and quantitative experiments validate the effectiveness of our approach, demonstrating that Stable-Hair v2 delivers results that approach commercial-level quality. By addressing the challenge of consistent hairstyle generation across views, our framework not only sets a new benchmark but also fills a critical gap in the domain of multi-view hairstyle modeling and transfer.

\bibliographystyle{IEEEtran}
\bibliography{main}

\begin{thebibliography}{10}
\providecommand{\url}[1]{#1}
\csname url@samestyle\endcsname
\providecommand{\newblock}{\relax}
\providecommand{\bibinfo}[2]{#2}
\providecommand{\BIBentrySTDinterwordspacing}{\spaceskip=0pt\relax}
\providecommand{\BIBentryALTinterwordstretchfactor}{4}
\providecommand{\BIBentryALTinterwordspacing}{\spaceskip=\fontdimen2\font plus
\BIBentryALTinterwordstretchfactor\fontdimen3\font minus \fontdimen4\font\relax}
\providecommand{\BIBforeignlanguage}[2]{{%
\expandafter\ifx\csname l@#1\endcsname\relax
\typeout{** WARNING: IEEEtran.bst: No hyphenation pattern has been}%
\typeout{** loaded for the language `#1'. Using the pattern for}%
\typeout{** the default language instead.}%
\else
\language=\csname l@#1\endcsname
\fi
#2}}
\providecommand{\BIBdecl}{\relax}
\BIBdecl

\bibitem{tan2020michigan}
Z.~Tan, M.~Chai, D.~Chen, J.~Liao, Q.~Chu, L.~Yuan, S.~Tulyakov, and N.~Yu, ``Michigan: Multi-input-conditioned hair image generation for portrait editing,'' \emph{ACM Transactions on Graphics (TOG)}, vol.~39, no.~4, pp. 1--13, 2020.

\bibitem{2018arXivhair-Gans}
M.~Zhang and Y.~Zheng, ``{Hair-GANs: Recovering 3D Hair Structure from a Single Image},'' \emph{arXiv e-prints}, p. arXiv:1811.06229, Nov. 2018.

\bibitem{2022CtrlHair}
X.~Guo, M.~Kan, T.~Chen, and S.~Shan, ``Gan with multivariate disentangling for controllable hair editing,'' \emph{ECCV}, 2022.

\bibitem{2022HairNet}
P.~Zhu, R.~Abdal, J.~Femiani, and P.~Wonka, ``Hairnet: Hairstyle transfer withpose changes,'' \emph{ECCV}, 2022.

\bibitem{2023HairNeRF}
S.~Chang, G.~Kim, and H.~Kim, ``Hairnerf: Geometry-aware image synthesis for hairstyle transfer,'' in \emph{Proceedings of the IEEE/CVF International Conference on Computer Vision}, 2023, pp. 2448--2458.

\bibitem{chung2022hairfit}
C.~Chung, T.~Kim, H.~Nam, S.~Choi, G.~Gu, S.~Park, and J.~Choo, ``Hairfit: pose-invariant hairstyle transfer via flow-based hair alignment and semantic-region-aware inpainting,'' \emph{BMVC}, 2021.

\bibitem{zhu2021barbershop}
P.~Zhu, R.~Abdal, J.~Femiani, and P.~Wonka, ``Barbershop: Gan-based image compositing using segmentation masks,'' 2021.

\bibitem{saha2021LOHO}
Saha, Rohit, Duke, Brendan, Shkurti, Florian, Taylor, Graham, Aarabi, and Parham, ``Loho: Latent optimization of hairstyles via orthogonalization,'' in \emph{CVPR}, 2021.

\bibitem{wei2022hairclip}
T.~Wei, D.~Chen, W.~Zhou, J.~Liao, Z.~Tan, L.~Yuan, W.~Zhang, and N.~Yu, ``Hairclip: Design your hair by text and reference image,'' \emph{Proceedings of the IEEE/CVF Conference on Computer Vision and Pattern Recognition}, 2022.

\bibitem{wei2023hairclipv2}
T.~Wei, D.~Chen, W.~Zhou, J.~Liao, W.~Zhang, G.~Hua, and N.~Yu, ``Hairclipv2: Unifying hair editing via proxy feature blending,'' \emph{ICCV}, 2023.

\bibitem{nikolaev2024hairfastgan}
M.~Nikolaev, M.~Kuznetsov, D.~Vetrov, and A.~Alanov, ``Hairfastgan: Realistic and robust hair transfer with a fast encoder-based approach,'' \emph{NeurIPS}, 2024.

\bibitem{Khwanmuang2023StyleGANSalon}
S.~Khwanmuang, P.~Phongthawee, P.~Sangkloy, and S.~Suwajanakorn, ``Stylegan salon: Multi-view latent optimization for pose-invariant hairstyle transfer,'' in \emph{IEEE Conference on Computer Vision and Pattern Recognition (CVPR)}, 2023.

\bibitem{kim2022styleSYH}
T.~Kim, C.~Chung, Y.~Kim, S.~Park, K.~Kim, and J.~Choo, ``Style your hair: Latent optimization for pose-invariant hairstyle transfer via local-style-aware hair alignment,'' \emph{ECCV}, 2022.

\bibitem{zhang2025stable}
Y.~Zhang, Q.~Zhang, Y.~Song, J.~Zhang, H.~Tang, and J.~Liu, ``Stable-hair: Real-world hair transfer via diffusion model,'' in \emph{Proceedings of the AAAI Conference on Artificial Intelligence}, vol.~39, no.~10, 2025, pp. 10\,348--10\,356.

\bibitem{chung2025hairfusion}
C.~Chung, S.~Park, J.~Kim, and J.~Choo, ``What to preserve and what to transfer: Faithful, identity-preserving diffusion-based hairstyle transfer,'' in \emph{The Association for the Advancement of Artificial Intelligence}, 2025.

\bibitem{Wu_2022_CVPR}
Y.~Wu, Y.-L. Yang, and X.~Jin, ``Hairmapper: Removing hair from portraits using gans,'' in \emph{Proceedings of the IEEE/CVF Conference on Computer Vision and Pattern Recognition (CVPR)}, June 2022, pp. 4227--4236.

\bibitem{voleti2024sv3d}
V.~Voleti, C.-H. Yao, M.~Boss, A.~Letts, D.~Pankratz, D.~Tochilkin, C.~Laforte, R.~Rombach, and V.~Jampani, ``{SV3D}: Novel multi-view synthesis and {3D} generation from a single image using latent video diffusion,'' in \emph{European Conference on Computer Vision (ECCV)}, 2024.

\bibitem{shu2022few}
C.~Shu, H.~Wu, H.~Zhou, J.~Liu, Z.~Hong, C.~Ding, J.~Han, J.~Liu, E.~Ding, and J.~Wang, ``Few-shot head swapping in the wild,'' in \emph{Proceedings of the IEEE/CVF Conference on Computer Vision and Pattern Recognition}, 2022, pp. 10\,789--10\,798.

\bibitem{chung2025preserve}
C.~Chung, S.~Park, J.~Kim, and J.~Choo, ``What to preserve and what to transfer: Faithful, identity-preserving diffusion-based hairstyle transfer,'' in \emph{Proceedings of the AAAI Conference on Artificial Intelligence}, vol.~39, no.~3, 2025, pp. 2582--2590.

\bibitem{ho2020denoising}
J.~Ho, A.~Jain, and P.~Abbeel, ``Denoising diffusion probabilistic models,'' \emph{arXiv preprint arxiv:2006.11239}, 2020.

\bibitem{dalle2}
A.~Ramesh, P.~Dhariwal, A.~Nichol, C.~Chu, and M.~Chen, ``Hierarchical text-conditional image generation with clip latents,'' 2022.

\bibitem{sdxl}
D.~Podell, Z.~English, K.~Lacey, A.~Blattmann, T.~Dockhorn, J.~M{\"u}ller, J.~Penna, and R.~Rombach, ``Sdxl: Improving latent diffusion models for high-resolution image synthesis,'' \emph{ICLR}, 2024.

\bibitem{Imagen}
C.~Saharia, W.~Chan, S.~Saxena, L.~Li, J.~Whang, E.~L. Denton, K.~Ghasemipour, R.~Gontijo~Lopes, B.~Karagol~Ayan, T.~Salimans \emph{et~al.}, ``Photorealistic text-to-image diffusion models with deep language understanding,'' \emph{Advances in neural information processing systems}, vol.~35, pp. 36\,479--36\,494, 2022.

\bibitem{rombach2022high}
R.~Rombach, A.~Blattmann, D.~Lorenz, P.~Esser, and B.~Ommer, ``High-resolution image synthesis with latent diffusion models,'' in \emph{Proceedings of the IEEE/CVF conference on computer vision and pattern recognition}, 2022, pp. 10\,684--10\,695.

\bibitem{IF}
C.~Saharia, W.~Chan, S.~Saxena, L.~Li, J.~Whang, E.~L. Denton, K.~Ghasemipour, R.~Gontijo~Lopes, B.~Karagol~Ayan, T.~Salimans \emph{et~al.}, ``Photorealistic text-to-image diffusion models with deep language understanding,'' \emph{Advances in Neural Information Processing Systems}, vol.~35, pp. 36\,479--36\,494, 2022.

\bibitem{attendandexcite}
H.~Chefer, Y.~Alaluf, Y.~Vinker, L.~Wolf, and D.~Cohen-Or, ``Attend-and-excite: Attention-based semantic guidance for text-to-image diffusion models,'' 2023.

\bibitem{layerdiffusion}
P.~Li, Q.~Huang, Y.~Ding, and Z.~Li, ``Layerdiffusion: Layered controlled image editing with diffusion models,'' \emph{SIGGRAPH Asia}, 2023.

\bibitem{selfguidance}
D.~Epstein, A.~Jabri, B.~Poole, A.~A. Efros, and A.~Holynski, ``Diffusion self-guidance for controllable image generation,'' \emph{NeurIPS}, 2023.

\bibitem{masactrl}
M.~Cao, X.~Wang, Z.~Qi, Y.~Shan, X.~Qie, and Y.~Zheng, ``Masactrl: Tuning-free mutual self-attention control for consistent image synthesis and editing,'' \emph{ICCV}, 2023.

\bibitem{dragondiffusion}
C.~Mou, X.~Wang, J.~Song, Y.~Shan, and J.~Zhang, ``Dragondiffusion: Enabling drag-style manipulation on diffusion models,'' \emph{ICLR}, 2023.

\bibitem{iedit}
R.~Bodur, E.~Gundogdu, B.~Bhattarai, T.-K. Kim, M.~Donoser, and L.~Bazzani, ``iedit: Localised text-guided image editing with weak supervision,'' \emph{CVPRW}, 2024.

\bibitem{sine}
Z.~Zhang, L.~Han, A.~Ghosh, D.~N. Metaxas, and J.~Ren, ``Sine: Single image editing with text-to-image diffusion models,'' in \emph{Proceedings of the IEEE/CVF Conference on Computer Vision and Pattern Recognition}, 2023, pp. 6027--6037.

\bibitem{tsaban2023ledits}
L.~Tsaban and A.~Passos, ``Ledits: Real image editing with ddpm inversion and semantic guidance,'' \emph{arXiv preprint arXiv:2307.00522}, 2023.

\bibitem{li2024zone}
S.~Li, B.~Zeng, Y.~Feng, S.~Gao, X.~Liu, J.~Liu, L.~Li, X.~Tang, Y.~Hu, J.~Liu \emph{et~al.}, ``Zone: Zero-shot instruction-guided local editing,'' in \emph{Proceedings of the IEEE/CVF Conference on Computer Vision and Pattern Recognition}, 2024, pp. 6254--6263.

\bibitem{controlnet}
L.~Zhang and M.~Agrawala, ``Adding conditional control to text-to-image diffusion models,'' \emph{ICCV}, 2023.

\bibitem{t2i}
C.~Mou, X.~Wang, L.~Xie, J.~Zhang, Z.~Qi, Y.~Shan, and X.~Qie, ``T2i-adapter: Learning adapters to dig out more controllable ability for text-to-image diffusion models,'' \emph{AAAI}, 2024.

\bibitem{unicontrol}
S.~Zhao, D.~Chen, Y.-C. Chen, J.~Bao, S.~Hao, L.~Yuan, and K.-Y.~K. Wong, ``Uni-controlnet: All-in-one control to text-to-image diffusion models,'' \emph{NeurIPS}, 2023.

\bibitem{directed}
W.-D.~K. Ma, J.~Lewis, W.~B. Kleijn, and T.~Leung, ``Directed diffusion: Direct control of object placement through attention guidance,'' \emph{arXiv preprint arXiv:2302.13153}, 2023.

\bibitem{zhang2024ssr}
Y.~Zhang, Y.~Song, J.~Liu, R.~Wang, J.~Yu, H.~Tang, H.~Li, X.~Tang, Y.~Hu, H.~Pan \emph{et~al.}, ``Ssr-encoder: Encoding selective subject representation for subject-driven generation,'' in \emph{Proceedings of the IEEE/CVF Conference on Computer Vision and Pattern Recognition}, 2024, pp. 8069--8078.

\bibitem{lora}
\BIBentryALTinterwordspacing
E.~J. Hu, Y.~Shen, P.~Wallis, Z.~Allen-Zhu, Y.~Li, S.~Wang, L.~Wang, and W.~Chen, ``Lo{RA}: Low-rank adaptation of large language models,'' in \emph{International Conference on Learning Representations}, 2022. [Online]. Available: \url{https://openreview.net/forum?id=nZeVKeeFYf9}
\BIBentrySTDinterwordspacing

\bibitem{domainagnostic}
M.~Arar, R.~Gal, Y.~Atzmon, G.~Chechik, D.~Cohen-Or, A.~Shamir, and A.~H. Bermano, ``Domain-agnostic tuning-encoder for fast personalization of text-to-image models,'' \emph{arXiv preprint arXiv:2307.06925}, 2023.

\bibitem{taming}
X.~Jia, Y.~Zhao, K.~C. Chan, Y.~Li, H.~Zhang, B.~Gong, T.~Hou, H.~Wang, and Y.-C. Su, ``Taming encoder for zero fine-tuning image customization with text-to-image diffusion models,'' \emph{arXiv preprint arXiv:2304.02642}, 2023.

\bibitem{TI}
\BIBentryALTinterwordspacing
R.~Gal, Y.~Alaluf, Y.~Atzmon, O.~Patashnik, A.~H. Bermano, G.~Chechik, and D.~Cohen-Or, ``An image is worth one word: Personalizing text-to-image generation using textual inversion,'' 2022. [Online]. Available: \url{https://arxiv.org/abs/2208.01618}
\BIBentrySTDinterwordspacing

\bibitem{hyperdreambooth}
N.~Ruiz, Y.~Li, V.~Jampani, W.~Wei, T.~Hou, Y.~Pritch, N.~Wadhwa, M.~Rubinstein, and K.~Aberman, ``Hyperdreambooth: Hypernetworks for fast personalization of text-to-image models,'' \emph{arXiv preprint arXiv:2307.06949}, 2023.

\bibitem{DB}
N.~Ruiz, Y.~Li, V.~Jampani, Y.~Pritch, M.~Rubinstein, and K.~Aberman, ``Dreambooth: Fine tuning text-to-image diffusion models for subject-driven generation,'' 2022.

\bibitem{zhang2024fast}
Y.~Zhang, Y.~Song, J.~Yu, H.~Pan, and Z.~Jing, ``Fast personalized text to image synthesis with attention injection,'' in \emph{ICASSP 2024-2024 IEEE International Conference on Acoustics, Speech and Signal Processing (ICASSP)}.\hskip 1em plus 0.5em minus 0.4em\relax IEEE, 2024, pp. 6195--6199.

\bibitem{xu2024ootdiffusion}
Y.~Xu, T.~Gu, W.~Chen, and C.~Chen, ``Ootdiffusion: Outfitting fusion based latent diffusion for controllable virtual try-on,'' \emph{AAAI}, 2025.

\bibitem{stablegarment}
R.~Wang, H.~Guo, J.~Liu, H.~Li, H.~Zhao, X.~Tang, Y.~Hu, H.~Tang, and P.~Li, ``Stablegarment: Garment-centric generation via stable diffusion,'' \emph{arXiv preprint arXiv:2403.10783}, 2024.

\bibitem{kim2023stableviton}
J.~Kim, G.~Gu, M.~Park, S.~Park, and J.~Choo, ``Stableviton: Learning semantic correspondence with latent diffusion model for virtual try-on,'' \emph{CVPR}, 2024.

\bibitem{zeng2023cat}
J.~Zeng, D.~Song, W.~Nie, H.~Tian, T.~Wang, and A.~Liu, ``Cat-dm: Controllable accelerated virtual try-on with diffusion model,'' \emph{CVPR}, 2024.

\bibitem{zhang2024stable}
Y.~Zhang, L.~Wei, Q.~Zhang, Y.~Song, J.~Liu, H.~Li, X.~Tang, Y.~Hu, and H.~Zhao, ``Stable-makeup: When real-world makeup transfer meets diffusion model,'' \emph{SIGGRAPH}, 2025.

\bibitem{peng2023implicit}
S.~Peng, C.~Geng, Y.~Zhang, Y.~Xu, Q.~Wang, Q.~Shuai, X.~Zhou, and H.~Bao, ``Implicit neural representations with structured latent codes for human body modeling,'' \emph{IEEE Transactions on Pattern Analysis and Machine Intelligence}, 2023.

\bibitem{peng2021neural}
S.~Peng, Y.~Zhang, Y.~Xu, Q.~Wang, Q.~Shuai, H.~Bao, and X.~Zhou, ``Neural body: Implicit neural representations with structured latent codes for novel view synthesis of dynamic humans,'' in \emph{CVPR}, 2021.

\bibitem{qian2024gaussianavatars}
S.~Qian, T.~Kirschstein, L.~Schoneveld, D.~Davoli, S.~Giebenhain, and M.~Nie{\ss}ner, ``Gaussianavatars: Photorealistic head avatars with rigged 3d gaussians,'' in \emph{Proceedings of the IEEE/CVF Conference on Computer Vision and Pattern Recognition}, 2024, pp. 20\,299--20\,309.

\bibitem{mildenhall2020nerf}
B.~Mildenhall, P.~P. Srinivasan, M.~Tancik, J.~T. Barron, R.~Ramamoorthi, and R.~Ng, ``Nerf: Representing scenes as neural radiance fields for view synthesis,'' in \emph{ECCV}, 2020.

\bibitem{kerbl3Dgaussians}
\BIBentryALTinterwordspacing
B.~Kerbl, G.~Kopanas, T.~Leimk{\"u}hler, and G.~Drettakis, ``3d gaussian splatting for real-time radiance field rendering,'' \emph{ACM Transactions on Graphics}, vol.~42, no.~4, July 2023. [Online]. Available: \url{https://repo-sam.inria.fr/fungraph/3d-gaussian-splatting/}
\BIBentrySTDinterwordspacing

\bibitem{liu2023zero}
R.~Liu, R.~Wu, B.~Van~Hoorick, P.~Tokmakov, S.~Zakharov, and C.~Vondrick, ``Zero-1-to-3: Zero-shot one image to 3d object,'' in \emph{Proceedings of the IEEE/CVF international conference on computer vision}, 2023, pp. 9298--9309.

\bibitem{shi2023mvdream}
Y.~Shi, P.~Wang, J.~Ye, M.~Long, K.~Li, and X.~Yang, ``Mvdream: Multi-view diffusion for 3d generation,'' \emph{ICLR}, 2024.

\bibitem{shi2023zero123++}
R.~Shi, H.~Chen, Z.~Zhang, M.~Liu, C.~Xu, X.~Wei, L.~Chen, C.~Zeng, and H.~Su, ``Zero123++: a single image to consistent multi-view diffusion base model,'' \emph{arXiv preprint arXiv:2310.15110}, 2023.

\bibitem{liu2023syncdreamer}
Y.~Liu, C.~Lin, Z.~Zeng, X.~Long, L.~Liu, T.~Komura, and W.~Wang, ``Syncdreamer: Generating multiview-consistent images from a single-view image,'' \emph{(ICLR)}, 2024.

\bibitem{deitke2023objaverse}
M.~Deitke, D.~Schwenk, J.~Salvador, L.~Weihs, O.~Michel, E.~VanderBilt, L.~Schmidt, K.~Ehsani, A.~Kembhavi, and A.~Farhadi, ``Objaverse: A universe of annotated 3d objects,'' in \emph{Proceedings of the IEEE/CVF conference on computer vision and pattern recognition}, 2023, pp. 13\,142--13\,153.

\bibitem{blattmann2023stable}
A.~Blattmann, T.~Dockhorn, S.~Kulal, D.~Mendelevitch, M.~Kilian, D.~Lorenz, Y.~Levi, Z.~English, V.~Voleti, A.~Letts \emph{et~al.}, ``Stable video diffusion: Scaling latent video diffusion models to large datasets,'' \emph{arXiv preprint arXiv:2311.15127}, 2023.

\bibitem{Chan2022}
E.~R. Chan, C.~Z. Lin, M.~A. Chan, K.~Nagano, B.~Pan, S.~D. Mello, O.~Gallo, L.~Guibas, J.~Tremblay, S.~Khamis, T.~Karras, and G.~Wetzstein, ``Efficient geometry-aware {3D} generative adversarial networks,'' in \emph{CVPR}, 2022.

\bibitem{xiang2023gramhd}
J.~Xiang, J.~Yang, Y.~Deng, and X.~Tong, ``Gram-hd: 3d-consistent image generation at high resolution with generative radiance manifolds,'' in \emph{Proceedings of the IEEE/CVF International Conference on Computer Vision (ICCV)}, October 2023, pp. 2195--2205.

\bibitem{ko20233d}
J.~Ko, K.~Cho, D.~Choi, K.~Ryoo, and S.~Kim, ``3d gan inversion with pose optimization,'' in \emph{Proceedings of the IEEE/CVF Winter Conference on Applications of Computer Vision}, 2023, pp. 2967--2976.

\bibitem{zhang2024flashface}
S.~Zhang, L.~Huang, X.~Chen, Y.~Zhang, Z.-F. Wu, Y.~Feng, W.~Wang, Y.~Shen, Y.~Liu, and P.~Luo, ``Flashface: Human image personalization with high-fidelity identity preservation,'' \emph{arXiv preprint arXiv:2403.17008}, 2024.

\bibitem{guo2023animatediff}
Y.~Guo, C.~Yang, A.~Rao, Z.~Liang, Y.~Wang, Y.~Qiao, M.~Agrawala, D.~Lin, and B.~Dai, ``Animatediff: Animate your personalized text-to-image diffusion models without specific tuning,'' \emph{International Conference on Learning Representations}, 2024.

\bibitem{fid}
M.~Heusel, H.~Ramsauer, T.~Unterthiner, B.~Nessler, and S.~Hochreiter, ``Gans trained by a two time-scale update rule converge to a local nash equilibrium,'' \emph{Advances in neural information processing systems}, vol.~30, 2017.

\bibitem{wang2004imageSSIM}
Z.~Wang, A.~C. Bovik, H.~R. Sheikh, and E.~P. Simoncelli, ``Image quality assessment: from error visibility to structural similarity,'' \emph{IEEE transactions on image processing}, vol.~13, no.~4, pp. 600--612, 2004.

\bibitem{arcface}
J.~Deng, J.~Guo, X.~Niannan, and S.~Zafeiriou, ``Arcface: Additive angular margin loss for deep face recognition,'' in \emph{CVPR}, 2019.

\bibitem{clip}
A.~Radford, J.~W. Kim, C.~Hallacy, A.~Ramesh, G.~Goh, S.~Agarwal, G.~Sastry, A.~Askell, P.~Mishkin, J.~Clark, G.~Krueger, and I.~Sutskever, ``Learning transferable visual models from natural language supervision,'' 2021.

\bibitem{deng2019accurate}
Y.~Deng, J.~Yang, S.~Xu, D.~Chen, Y.~Jia, and X.~Tong, ``Accurate 3d face reconstruction with weakly-supervised learning: From single image to image set,'' in \emph{Proceedings of the IEEE/CVF conference on computer vision and pattern recognition workshops}, 2019, pp. 0--0.

\end{thebibliography}

\end{document}